\documentclass[conference,compsoc]{IEEEtran}
\ifCLASSOPTIONcompsoc
  \usepackage[nocompress]{cite}
\else
  \usepackage{cite}
\fi
\ifCLASSINFOpdf
\else
\fi

\usepackage{enumitem}
\usepackage{multirow}
\usepackage{subcaption}
\usepackage{tikz}
\usepackage{booktabs}
\usepackage{bm}
\usepackage{wasysym}
\usepackage{pifont}
\usepackage{soul}
\usepackage[table,xcdraw]{xcolor}
\usepackage[hidelinks]{hyperref}

\newcommand{\scheme}{BVBench}
\newcommand{\cmark}{\ding{51}}
\newcommand{\xmark}{\ding{55}}
\newcommand{\fstar}{\ding{72}}
\newcommand{\hstar}{\ding{73}}

\begin{document}
\title{Understanding Backdoor Vulnerabilities in Vertical Federated Learning:\\The Gap Between Research and Practice}

\author{
Ziqi Zhao,
Jialin Lu,
Junjie Shan,
Junyuan Zhang,
Shuya Yang,
Ka-Ho Chow
\\
School of Computing and Data Science, The University of Hong Kong, Hong Kong SAR, China
}

\maketitle

\begin{abstract}
	Vertical Federated Learning (VFL) enables organizations holding complementary features of shared entities to collaborate and train models. In this setting, the initiator can withhold information about the learning task, while other contributors participate without exposing their local datasets, creating an asymmetric information structure aligned with growing privacy demands. However, this asymmetry is a double-edged sword. Among various threats, backdoor attacks are particularly concerning because VFL not only enables malicious contributors to poison the model during training, but also allows them to activate the backdoor at inference time to manipulate predictions. Although prior work has reported near-perfect attack success rates and proposed effective defenses, we find that most findings fail to hold under realistic conditions, exposing a fundamental gap between research and practice. In this paper, we present a systematic, practice-oriented study of backdoor vulnerabilities in VFL, revealing this gap in both methodological design and evaluation practices. We show that existing approaches overlook key practical constraints and therefore rely on unrealistic prior knowledge. Furthermore, these limitations have remained hidden due to poorly designed evaluation practices in the literature. To bridge this gap, we redefine threat models under realistic constraints, propose practical backdoor workflows, and introduce \scheme{}, a backdoor-centric benchmark that enables fair, practical, and comprehensive evaluation, preloaded with state-of-the-art baselines. \scheme{} provides strong evidence of the fragility of the current understanding of VFL backdoor risks and establishes a foundation for steering research toward uncovering practical vulnerabilities and developing more meaningful defenses.
\end{abstract}

\IEEEpeerreviewmaketitle

\section{Introduction}
Cross-organizational collaboration is key to unlocking the full potential of modern machine learning (ML), which thrives on rich and expressive data to achieve high predictive performance. In practice, organizations across domains often possess complementary features about the same set of entities, such as credit history in financial institutions~\cite{zheng2020verticalfederatedlearningcredit} and health conditions in healthcare providers~\cite{vepakomma2018split,huang2023verticalfederatedknowledgehealthcare}. Integrating these distributed features can enhance data expressiveness and improve predictive performance. However, despite enabling multi-billion-dollar opportunities, such collaborations remain difficult due to data privacy regulations~\cite{EuropeanParliament2016a, ccpa2018} that prohibit sharing raw data across organizational boundaries.

\begin{figure}
	\centering
	\includegraphics[width=0.99\linewidth]{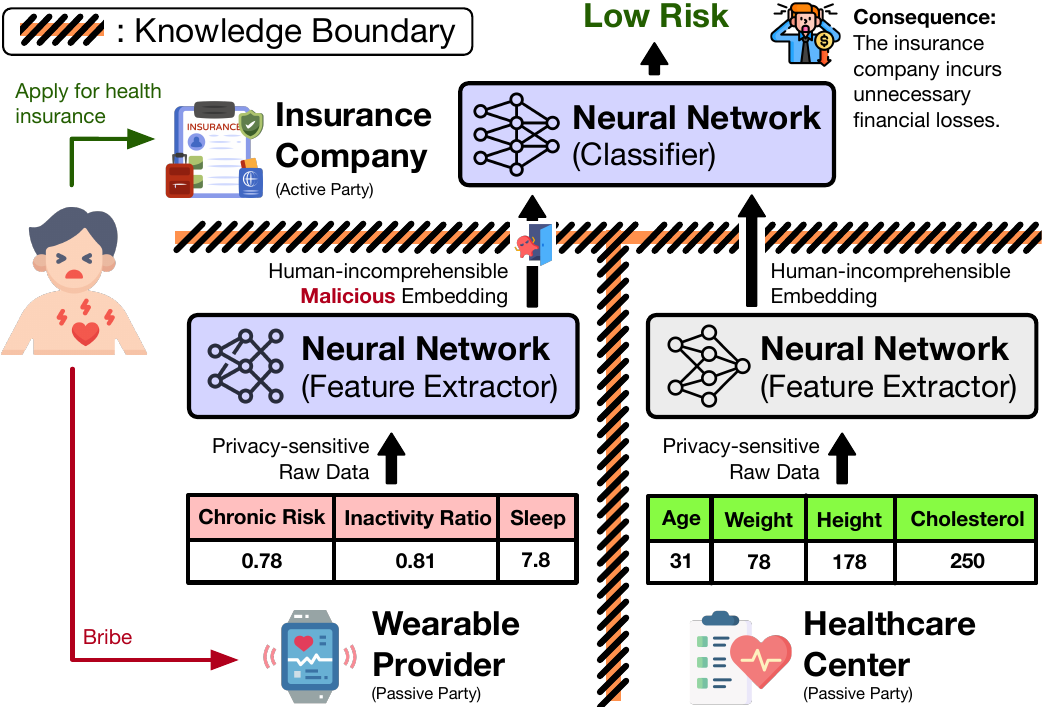}
	\caption{VFL enables an active party (e.g., an insurance company) to collaborate with passive parties (e.g., a wearable provider and a healthcare center) and leverage complementary features of shared entities to improve ML tasks, such as underwriting risk assessment, without sharing raw data. However, the inherent information asymmetry in VFL can be exploited: a malicious participant (e.g., the wearable provider) may implant a backdoor during training and later manipulate model predictions at inference time.}\label{fig:teaser}
\end{figure}

Vertical Federated Learning (VFL)~\cite{vepakomma2018split, liuVerticalFederatedLearning2024,yeVerticalFederatedLearning2025a,khanVerticalFederatedLearning2025} has emerged as a promising solution to this challenge. As illustrated in Figure~\ref{fig:teaser}, an insurance company (the ``active'' party) can collaborate with a wearable provider and a healthcare center (the ``passive'' parties). When a potential client applies for insurance, the active party requests the passive parties to identify matching records, process them locally through private feature extractors, and return the resulting embeddings. 
These embeddings, which are not human-interpretable, are aggregated by the active party and used by a classifier to predict underwriting risk. In this way, predictions can leverage sensitive information, such as health status and activity data, without exposing raw data. This capability is enabled by jointly training models across parties while preserving data locality and creating information asymmetry: the active party knows the ML task but not the raw data, whereas passive parties know their local data but not the task itself. While this asymmetry makes cross-organizational learning practical, it also introduces unique security risks.

Backdoor attacks~\cite{xuanPracticalGeneralBackdoor2023, baiVILLAINBackdoorAttacks2023,liuTargetedPoisoningAttacks2025} are a particularly concerning threat in VFL. In such attacks, an adversary manipulates the training process to implant a hidden association between a trigger pattern and a target prediction. At inference time, the presence of the trigger causes the model to produce adversary-chosen outputs while behaving normally otherwise. VFL provides a unique and favorable setting for such attacks: passive parties participate in both training, where backdoors can be implanted, and inference, where they can be activated. For example, in Figure~\ref{fig:teaser}, a malicious wearable provider could implant a backdoor during training and later inject trigger patterns into embeddings at inference time, overriding the classifier’s decision (e.g., approving an otherwise ineligible insurance application). Such scenarios raise serious concerns about the security of VFL systems and have motivated a growing body of research on attack and defense mechanisms. However, do these studies truly reflect how VFL backdoors behave in real-world deployments?

In this paper, we argue that the current understanding of backdoor vulnerabilities in VFL is fragile. Despite rapid research progress, it remains unclear to what extent existing findings generalize to practice. This gap has important consequences: practitioners may underestimate risks when systems appear robust, or overtrust defenses that fail under realistic conditions. By systematizing existing work, we identify a disconnect between how VFL backdoor vulnerabilities are studied and how VFL systems operate in practice. We attribute this gap to misalignment along two dimensions: (i) methodological design and (ii) evaluation design.

\begin{figure}
	\includegraphics[width=\linewidth]{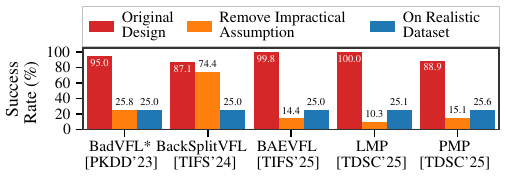}
	\caption{The high attack success rates (red) reported in prior work can be misleading. Existing approaches often rely on assumptions that are unlikely to hold in practice (orange) or are evaluated under unrealistic experimental settings (blue).}\label{fig:kf-evidence}
\end{figure}

\textbf{Methodological Design.} Existing attacks and defenses often rely on unrealistic assumptions regarding threat models and, consequently, problem formulation. For example, many attacks assume access to seemingly obtainable knowledge of the ML task (e.g., class semantics), even though the active party has no obligation to share such information. In practice, a malicious passive party is unlikely to know the task to which it contributes, including the number of classes, let alone their meanings. As shown in Figure~\ref{fig:kf-evidence} (orange), relaxing this assumption significantly reduces the success rate of state-of-the-art attacks. This finding suggests that prior work may fail to generalize to real deployments and overestimate attack effectiveness, highlighting the need to ground future research in more realistic assumptions.

\textbf{Evaluation Design.} The evaluation of VFL backdoor vulnerabilities is fragmented. Existing studies rely on unrealistic datasets, inconsistent experimental setups, and limited metrics that fail to capture critical properties. More importantly, attack and defense performance is highly sensitive to these choices. As shown in Figure~\ref{fig:kf-evidence} (blue), attacks reported to achieve near-perfect success rates can only reach a meaningless level of effectiveness on realistic VFL datasets, an effect previously overlooked due to the lack of standardized benchmarks and reproducible implementations. Overall, current evaluations do not provide a reliable or objective understanding of practical security risks.

As interest in VFL continues to grow in both academia and industry, the absence of a practice-oriented framework risks fostering misconceptions and hindering progress. To address this gap, we make the following contributions:
\begin{itemize}[leftmargin=*, noitemsep, topsep=0pt]
	\item \textbf{Knowledge Systematization.} We provide a structured analysis of backdoor attacks and defenses, uncover their limitations, redefine a threat model grounded in realistic assumptions, and identify a practical backdoor workflow.
	\item \textbf{Evaluation Framework.} We develop \scheme{}, a VFL backdoor-centric benchmark that includes realistic datasets, comprehensive metrics, and a standardized evaluation recipe for fair and reproducible evaluations.
	\item \textbf{Empirical Studies.} We evaluate existing methods under \scheme{} and show that they exhibit limited practicality and are highly sensitive to evaluation choices.
\end{itemize}
Our work benefits multiple stakeholders. For researchers, it clarifies realistic threat models and highlights open challenges under a practical backdoor workflow. For practitioners, it provides new understanding of risks and offers tools and guidelines for rigorous evaluation. Our artifacts are \href{https://github.com/HKU-TASR/BVBench}{here}.

\section{Background}\label{sec:background}
\subsection{Vertical Federated Learning}
Although VFL supports various model families, neural network-based VFL~\cite{liuLabelLeakageVertical2024, yanCrossmodalVerticalFederated2024, shenMARSVFLUnifiedBenchmark2025, weiFedAdsBenchmarkPrivacyPreserving2023, wangUnifiedSolutionPrivacy2024, ADIAdversarialDominating2023, yuanGeneralTestTimeBackdoor2025} has been the primary focus of recent research due to its strong representational power. We therefore focus on this setting. Figure~\ref{fig:vfl-internals} illustrates a system with one active party and $N$ passive parties holding complementary features for shared entities. VFL consists of three stages:
\begin{figure}
	\centering
	\includegraphics[width=0.96\linewidth]{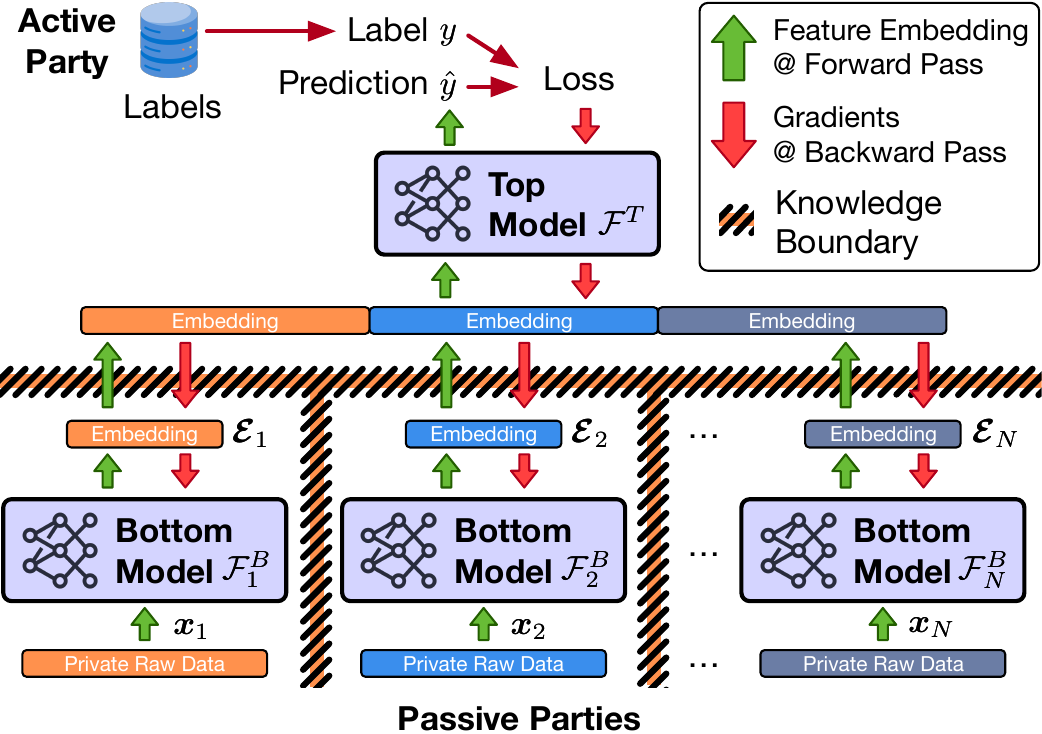}
	\caption{Information asymmetry is fundamental to VFL. During the forward pass (green), passive parties send only embeddings to the active party, which aggregates them to produce predictions. During the backward pass (red), the active party returns gradients with respect to each party’s embeddings. As a result, the active party has access to the task and labels but not the raw data, while passive parties retain raw data but have no visibility into the task or labels.}\label{fig:vfl-internals}
	\vspace{-1em}
\end{figure}

\textbf{Stage 1: Negotiation.} The active party defines the ML task and collects labels for existing entities to be used for training and evaluation. Because it may lack sufficient features, or even any features at all, the active party recruits passive parties to contribute complementary data. The parties negotiate incentives and agree on system design choices (e.g., model architectures and optimization strategies), as well as participation in both training and inference.

\textbf{Stage 2: Training.} Training begins with privacy-preserving entity alignment across parties, such as Private Set Intersection (PSI)~\cite{pinkas2018scalable, yangOpenVFLVerticalFederated2024b,xiPrivateSampleAlignment2025}. At each iteration, the active party selects a minibatch of sample IDs and distributes them to passive parties (we use batch size $1$ for brevity).
\begin{itemize}[leftmargin=*, noitemsep, topsep=0pt]
	\item \emph{Forward Pass}: As shown in Figure~\ref{fig:vfl-internals} (green path), each passive party $i$ retrieves its local data $\boldsymbol{x}_i$ and feeds it into its bottom model $\mathcal{F}^B_i$ to produce an embedding $\boldsymbol{\mathcal{E}}_i$, which is then sent to the active party. The active party concatenates these embeddings $\boldsymbol{\mathcal{E}}_1 \oplus \cdots \oplus \boldsymbol{\mathcal{E}}_N$ and applies the top model $\mathcal{F}^T$ as a classifier to generate prediction $\hat{y}$.
	\item \emph{Backward Pass}: As shown in Figure~\ref{fig:vfl-internals} (red path), the active party computes the loss using predicted label $\hat{y}$ and ground-truth label $y$, and updates the top model $\mathcal{F}^T$ via gradient descent. By the chain rule, it then computes and sends gradients w.r.t. each $\boldsymbol{\mathcal{E}}_i$ to the corresponding passive party $i$, which continues backpropagation to update its bottom model $\mathcal{F}^B_i$. Overall, all models ($\mathcal{F}^T, \mathcal{F}^B_1, ..., \mathcal{F}^B_N$) are jointly optimized to improve prediction accuracy.
\end{itemize}
This process continues until convergence or until a predefined number of epochs is reached.

\textbf{Stage 3: Inference.} In VFL, all parties must participate during inference. Upon receiving a query, the active party broadcasts the sample ID, requests embeddings from passive parties, and performs the forward pass described above.

\subsection{Backdoor Vulnerabilities}

\textbf{Attacks.} 
Backdoor attacks manipulate model behavior by implanting a hidden trigger during training that induces a target prediction at inference time. This is achieved by injecting a trigger pattern into a subset of training samples, allowing the model to associate the trigger with the desired output. A backdoor attack aims to preserve utility on clean inputs, induce the target output when the trigger is present, and remain difficult to detect during both the training-time manipulation and inference-time injection. While backdoor attacks have been extensively studied in centralized learning and horizontal federated learning (HFL)~\cite{gu2017badnets, chen2017targeted}, where clients hold disjoint samples with shared features, VFL introduces distinct opportunities and challenges.
\begin{itemize}[leftmargin=*, noitemsep, topsep=0pt]
	\item \emph{Opportunities}: VFL requires passive parties to participate in training and inference and to communicate through embeddings, creating two advantages for the adversary $i$. (i) Triggers can be crafted and injected directly in the embedding space (i.e., $\boldsymbol{\mathcal{E}}_i$) rather than the input space (i.e., $\boldsymbol{x}_i$). This avoids input-space constraints (e.g., valid image pixel ranges). (ii) A malicious passive party has consistent control over part of the top model input, eliminating the need for extra mechanisms to inject triggers.
	\item \emph{Challenges}: Passive parties cannot observe ground-truth labels~\cite{fuLabelInferenceAttacks2022,gaoLabelPrivacySource2024b}. This complicates establishing the association between triggers and target outputs, which in non-VFL settings is typically achieved by consistently injecting triggers into training samples of the target class and allowing the model to learn this shortcut. In addition, only a small fraction of the entire model is accessible to the adversary (i.e., its own bottom model $\mathcal{F}^B_i$), making it harder to control both attack effectiveness and stealth.
\end{itemize}

\textbf{Defenses.}
Backdoor defenses aim to preserve model utility while preventing trigger-induced misbehavior. In VFL, defenses are the responsibility of the active party, which orchestrates the system and produces predictions for downstream use. Although a wide range of defenses has been proposed in centralized and HFL settings~\cite{wang2019neural, tran2018spectral,liu2018fine, nguyenFlameTamingBackdoors2022, fungLimitationsFederatedLearning2020}, VFL presents distinct opportunities and challenges.
\begin{itemize}[leftmargin=*, noitemsep, topsep=0pt]
	\item \emph{Opportunities}: The active party observes embeddings from all passive parties ($\boldsymbol{\mathcal{E}}_1, ..., \boldsymbol{\mathcal{E}}_N$), while each passive party $i$ only observes its own contributions (i.e., $\boldsymbol{\mathcal{E}}_i$). This asymmetry enables cross-party consistency checks and the potential identification of anomalous embeddings.
	\item \emph{Challenges}: Because all parties are required during inference, excluding a suspected malicious party is impractical, as the entire system typically needs to be reconstructed from scratch. Hence, unlike non-VFL settings, mitigating backdoor effects is more preferred than detection alone.
\end{itemize}

In summary, VFL introduces unique structural constraints that fundamentally shape the design of attacks and defenses. Next, we systematically analyze existing approaches (Sections~\ref{sec:attack} and~\ref{sec:defense}) and identify key limitations in current evaluation practices, motivating the need for a comprehensive benchmark (Section~\ref{sec:benchmark}).

\begin{table*}
	\centering
	\setlength\tabcolsep{2.pt}
	\caption{Current VFL backdoor attacks exhibit practicality gaps in threat model and their current designs are either at odds with or poorly aligned with the practical backdoor workflow outlined in Section~\ref{sec:attack-design}.}
	\label{tab:tax-attack}
	\begin{tabular*}{\textwidth}{@{\extracolsep{\fill}}l|lc|lc|llc|lc@{}}
		\toprule
		\multirow[c]{4}{*}{\textbf{Attack}} & \multicolumn{2}{c|}{\multirow{2}{*}[-0.5ex]{\textbf{Threat Model}}} & \multicolumn{5}{c|}{\textbf{Attack Design}} & \multicolumn{2}{c}{\multirow{2}{*}[-0.5ex]{\textbf{\begin{tabular}[c]{@{}c@{}}Stealthiness\end{tabular}}}} \\
		\cmidrule(r){4-8}
		& & & \multicolumn{2}{c|}{\scriptsize\textbf{Label Acquisition}} & \multicolumn{3}{c|}{\scriptsize\textbf{Trigger Design for Multi-Target Learning}} & & \\
		\cmidrule(r){2-3}\cmidrule(r){4-5}\cmidrule(r){6-8} \cmidrule(r){9-10}
		& \multicolumn{1}{c}{{\scriptsize\textbf{Knowledge}}} & \scriptsize\textbf{Accessible?} & \multicolumn{1}{c}{\scriptsize\textbf{Signal}} & {\scriptsize\textbf{Feasible?}} & {\scriptsize\textbf{Trigger Space}} & {\scriptsize\textbf{Trigger Pattern}} & {\scriptsize\textbf{Scalable?}} & {\scriptsize\textbf{Constraint}} & \scriptsize\textbf{Undetectable?}  \\ \midrule
		BackSplitVFL~\cite{heBackdoorAttackSplit2024} & $\mathcal{L}_{target}$ & \xmark & - & - & Embedding & Class-aware & \LEFTcircle & $\ell_\infty$-norm & \LEFTcircle  \\
		LMP~\cite{liuTargetedPoisoningAttacks2025} & $\mathcal{L}_{target}$ & \xmark & - & - & Embedding & Class-aware & \LEFTcircle & - & \Circle  \\
		PMP~\cite{liuTargetedPoisoningAttacks2025} & $\mathcal{L}_{target}$ & \xmark & - & - & Embedding & Class-aware & \LEFTcircle & - & \Circle  \\
		VILLAIN~\cite{baiVILLAINBackdoorAttacks2023} & $\mathcal{L}_{target}$ & \xmark & Gradients & \LEFTcircle & Embedding & Fixed & \Circle & Std. Dev. & \LEFTcircle  \\
		BadVFL*~\cite{xuanPracticalGeneralBackdoor2023} & $\mathcal{L}_{target}$ & \xmark & Gradients & \LEFTcircle & Input & Fixed & \Circle & - & \Circle  \\
		LFBA~\cite{shenLabelFreeBackdoorAttacks2025} & $\mathcal{L}_{target}$ & \xmark & Gradients & \LEFTcircle & Input & Fixed & \Circle & - & \Circle  \\
		BadVFL~\cite{naseriBadVFLBackdoorAttacks2024} & $\mathcal{L}_{full}$ & \xmark & Embeddings & \LEFTcircle & Input & Class-aware & \LEFTcircle & - & \Circle  \\
		HijackVFL ($\mathcal{P}$, $\times$)~\cite{qiuHijackVerticalFederated2024} & $\mathcal{P}$ & \xmark & Embeddings & \LEFTcircle & Embedding & Class-aware & \LEFTcircle & - & \Circle  \\
		HijackVFL ($\mathcal{P}$, $\mathcal{L}_{target}$)~\cite{qiuHijackVerticalFederated2024} & $\mathcal{P}$+$\mathcal{L}_{target}$ & \xmark & Embeddings & \LEFTcircle & Embedding & Class-aware & \LEFTcircle & - & \Circle  \\
		HijackVFL ($\times$, $\mathcal{L}_{full}$)~\cite{qiuHijackVerticalFederated2024} & $\mathcal{L}_{full}$ & \xmark & Embeddings & \LEFTcircle & Embedding & Class-aware & \LEFTcircle & - & \Circle  \\
		BAEVFL~\cite{chenBackdoorAttackEncryptionProtected2025} & $\mathcal{L}_{target}$ & \xmark & Embeddings & \LEFTcircle & Input& Class-aware & \LEFTcircle & - & \Circle  \\ \bottomrule
	\end{tabular*}
\end{table*}
\section{Systematizing VFL Backdoor Attacks}\label{sec:attack}
Current backdoor attacks share a similar workflow. First, they collect or infer labels for some training samples. Then, they craft a trigger pattern and use those labeled samples to establish a shortcut between this trigger and the target class. We systematize existing works through the lens of practicality in the threat model, attack design, and stealthiness.

\subsection{Threat Model}\label{sec:threat-model}
Threat models define the conditions under which an attack is launched. Ensuring practicality is critical, as it determines whether the insights reflect real-world vulnerabilities. We summarize existing threat models as follows:

\subsubsection{Attacker's Goal}\label{sec:tm-goal} 
The adversary (a passive party $i$) constructs a trigger injection function $\mathcal{I}(\mathcal{F}_i^B, \boldsymbol{x}_i, t)$ for a designated target class $t$ such that, when activated during inference, the prediction by the top model $\mathcal{F}^T(\boldsymbol{\mathcal{E}}_1 \oplus \cdots \oplus \mathcal{I}(\mathcal{F}_i^B, \boldsymbol{x}_i, t) \oplus \cdots \oplus \boldsymbol{\mathcal{E}}_N)=t$. When the adversary submits benign embeddings produced by its bottom model $\mathcal{F}^B_i$ (i.e., $\mathcal{F}^B_i(\boldsymbol{x}_i)$), the prediction should maintain normal accuracy. This behavior is achieved by poisoning the VFL training process while remaining undetectable by the active party.

\subsubsection{Attacker's Capabilities}\label{tm-cap} 
Most existing attacks assume capabilities consistent with those of a passive party in standard VFL. During training, the adversary can submit arbitrary embeddings to the active party and manipulate received gradients to update its bottom model. During inference, the adversary can submit trigger-injected embeddings to activate the backdoor and influence predictions.

\subsubsection{Attacker's Knowledge} 
The main distinction across existing attacks lies in their assumptions about prior knowledge, as summarized in Table~\ref{tab:tax-attack} (Columns 2--3). The purpose of prior knowledge is to help acquire labels of training samples, which are essential for the attacker to craft triggers and establish the connection between the trigger and the target class. Current works make various assumptions, including:
\begin{itemize}[leftmargin=*, noitemsep, topsep=0pt]
	\item \textbf{Target-Class Label $\mathcal{L}_{{target}}$.} All attacks require the adversary to be able to identify or infer some training samples belonging to the target class. For example, VILLAIN~\cite{baiVILLAINBackdoorAttacks2023} and BadVFL$^*$~\cite{xuanPracticalGeneralBackdoor2023} require one such sample, while BackSplitVFL~\cite{heBackdoorAttackSplit2024} and LMP~\cite{liuTargetedPoisoningAttacks2025} require multiple. This requirement is unlikely to be met because the active party is not obligated to disclose task-specific information. Passive participants do not even know about the semantics of the classes, let alone identify a target class for attack. 
	\item \textbf{Full Label Space $\mathcal{L}_{full}$.} Some attacks, such as BadVFL~\cite{naseriBadVFLBackdoorAttacks2024}, further assume knowledge of the entire label space, including labeled samples for all classes or the total number of classes. For the same reason, this is even more unlikely to be accessible than the target-class label alone.
	\item \textbf{Top Model Posterior $\mathcal{P}$.} Some attacks assume the adversary can observe the top model's posterior during inference, which will be used to infer labels and optimize the trigger. This is in direct conflict with the VFL setting where the passive parties have no access to the top model.
\end{itemize}

\noindent In short, the information asymmetry in VFL makes such task-related knowledge inaccessible. We therefore redefine a practical threat model for VFL backdoor attacks as follows:

\begin{figure}[h!]\centering\vspace{-0.7em}
	\begin{tikzpicture}
		\node[draw, fill=gray!10, rectangle, rounded corners, inner sep=4pt] (box) {
			\begin{minipage}{0.97\linewidth}
				\textbf{Practical Threat Model:} The attacker possessing the capabilities as a passive party (\S\ref{tm-cap}) should achieve the attack goal (\S\ref{sec:tm-goal}) without prior knowledge of the ML task, such as class semantics, target-class labels, full label space, or any other task-related information.
			\end{minipage}
		};
	\end{tikzpicture}\vspace{-0.7em}
\end{figure}

\noindent Under this threat model, no existing attacks remain feasible. 

\subsection{Attack Design}\label{sec:attack-design}
Based on the redefined threat model, we outline a practical backdoor workflow as follows:

\begin{figure}[h!]\centering\vspace{-0.7em}
	\begin{tikzpicture}
		\node[draw, fill=gray!10, rectangle, rounded corners, inner sep=4pt] (box) {
			\begin{minipage}{0.97\linewidth}
				\textbf{Practical Backdoor Workflow:} First, label acquisition should aim to cluster training samples of the same class without access to task-related knowledge. Second, each cluster should be assigned a pseudo-label and a dedicated trigger pattern, and the class semantics of each pseudo-label should be inferred after deployment by sending trigger-injected queries and observing responses. Third, backdoor learning should be formulated as a multi-target instance while balancing effectiveness across targets. 
			\end{minipage}
		};
	\end{tikzpicture}\vspace{-0.7em}
\end{figure}

\noindent Next, we show that current attack designs fall short in aligning with this workflow.

\subsubsection{Label Acquisition} 
\begin{figure}[t]
	\centering
	\begin{subfigure}[b]{\linewidth}
        \centering
        \includegraphics[width=\linewidth]{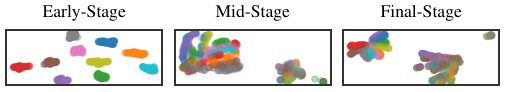}
        \caption{Gradients}
    \end{subfigure}\vspace{0.5em}
	\begin{subfigure}[b]{\linewidth}
        \centering
        \includegraphics[width=\linewidth]{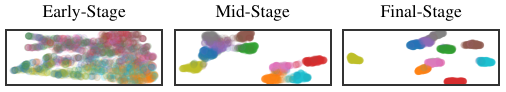}
        \caption{Embeddings}
    \end{subfigure}%
	\caption{Label inference can be unstable. The cluster effect of gradients and embeddings varies across training progress.}
	\label{fig:lia}
\end{figure}

Existing attacks rely on inaccessible labeled samples. Most of them justify this assumption by requiring only a few such samples and inferring the rest. They either exploit the gradients from the active party or the embeddings produced by the bottom model (Columns 4--5, Table~\ref{tab:tax-attack}), but such signals are unstable across training. As shown in Figure~\ref{fig:lia}, the former is informative only in early training, while the latter remains useless until the bottom model begins to extract useful embeddings. In practice, learning progress depends on uncontrollable factors such as task complexity. As analyzed in Section~\ref{sec:experiments-attack}, label quality plays a decisive role in backdoor attacks. Robust label acquisition from these unstable signals remains overlooked.

\subsubsection{Trigger Design for Multi-Target Learning}
\begin{figure}
	\centering
	\includegraphics[width=\linewidth]{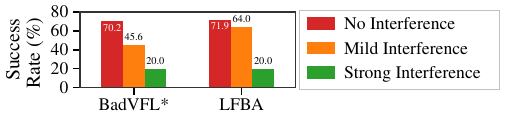}
	\caption{Trigger interference can hinder attack effectiveness in multi-target settings.}
	\label{fig:trigger-conflict}
\end{figure}

Multi-target learning poses a significant challenge to trigger design for three reasons (Columns 6--8, Table~\ref{tab:tax-attack}). First, each pseudo-label requires a distinct trigger pattern and/or position, while some existing methods (VILLAIN~\cite{baiVILLAINBackdoorAttacks2023}, BadVFL$^*$~\cite{xuanPracticalGeneralBackdoor2023} and LFBA~\cite{shenLabelFreeBackdoorAttacks2025}) use fixed and class-agnostic triggers, so the trigger pattern and injected position are the same for any target class. Second, it is non-trivial to scale to multiple targets even if trigger designs are class-aware because of potential interference. As shown in Figure~\ref{fig:trigger-conflict}, the coexistence of certain triggers can lead to mild (orange) or even drastic (green) degradation in attack effectiveness. Third, as more shortcuts need to be memorized, the trigger needs to produce stronger signals for the top model to discover. While our empirical analysis found that embedding-space triggers can be more effective, current methods provide no mechanism for balancing the attack strength across different classes, which can lead to cross-target conflicts and instability. 

\subsection{Stealthiness Consideration}\label{sec:embedding-space-stealthiness}
\begin{figure}
	\centering
	\includegraphics[width=\linewidth]{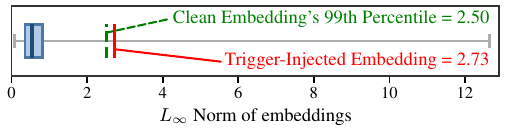} 
	\caption{Existing stealthiness constraints do not help attacks remain under the radar. For instance, BackSplitVFL~\cite{heBackdoorAttackSplit2024} constrains trigger norm to be within the maximum norm of clean embeddings, but it is still highly anomalous and can be easily detected using the 99\% quantile.}
	\label{fig:demo-stealthiness}
\end{figure}

Current stealthiness definitions are oversimplified and under-defined. Existing definitions rely on simple heuristics, such as repetitive patterns in the embedding space~\cite{heBackdoorAttackSplit2024}, or an abnormal range in the input or embedding space~\cite{heBackdoorAttackSplit2024,baiVILLAINBackdoorAttacks2023}. We argue that stealthiness should be defined from the defender's perspective and with the defender's knowledge, while remaining assessable under the attacker's knowledge and capabilities. Since embeddings are used in both training and inference and are accessible to both active and passive parties, they are the natural medium for defining stealthiness: if poisoned samples are easily identifiable in the embedding space, then the attack is not stealthy. Based on this intuition, we find that only BackSplitVFL~\cite{heBackdoorAttackSplit2024} and VILLAIN~\cite{baiVILLAINBackdoorAttacks2023} have relevant embedding-space constraints (Columns 9--10, Table~\ref{tab:tax-attack}). That said, these constraints cannot hide trigger-injected embeddings from defenders. For example, as shown in Figure~\ref{fig:demo-stealthiness}, BackSplitVFL uses maximum per-dimension and overall norm of clean embeddings to constrain poisoned embeddings, but such a value is already highly anomalous; VILLAIN uses a trigger pattern based on the averaged standard deviation over dimensions that have high standard deviation, but adding such triggers can still lead to off-distribution embeddings. Furthermore, the defender can access all parties' embeddings, which can make the attack even easier to detect if the defender leverages cross-party consistency checks.

\begin{figure}[h!]\centering\vspace{-0.7em}
	\begin{tikzpicture}
		\node[draw, fill=gray!10, rectangle, rounded corners, inner sep=4pt] (box) {
			\begin{minipage}{0.97\linewidth}
				\textbf{Practicality Insights:} Current stealthiness definitions are fragile and poorly aligned with defender's knowledge. It should be defender-driven, attacker-assessible, defined in the embedding space, and tight enough to make malicious embeddings indistinguishable from benign ones.
			\end{minipage}
		};
	\end{tikzpicture}\vspace{-1em}
\end{figure}

\section{Systematizing VFL Backdoor Defenses}\label{sec:defense}

\begin{table}
	\centering
	\setlength\tabcolsep{.7pt}
	\caption{Current VFL backdoor defenses exhibit practicality gaps in prior knowledge, design, and recovery quality.}
	\label{tab:tax-defense}
	\begin{tabular}{@{\extracolsep{\fill}}l|cc|ccc|c@{}}
		\toprule
		\multirow{3}{*}{\textbf{Defense}} & \multicolumn{2}{c|}{\textbf{Knowledge}} & \multicolumn{3}{c|}{\textbf{Defense Design}} & \multirow{3}{*}{\centering\textbf{\begin{tabular}[c]{@{}c@{}}Utility\\Recovery\end{tabular}}\par} \\
		\cmidrule(lr){2-3}\cmidrule(lr){4-6}
			& \multirow{2}{*}{\footnotesize\textbf{\begin{tabular}[c]{@{}c@{}}Ref.-\\Free\end{tabular}}} & \multirow{2}{*}{\footnotesize\textbf{\begin{tabular}[c]{@{}c@{}}Adv. Kn.-\\Free\end{tabular}}} & \multirow{2}{*}{\footnotesize\textbf{\begin{tabular}[c]{@{}c@{}}Pro-\\active?\end{tabular}}} & \multirow{2}{*}{\footnotesize\textbf{\begin{tabular}[c]{@{}c@{}}Multi-\\Target?\end{tabular}}} & \multirow{2}{*}{\footnotesize\textbf{\begin{tabular}[c]{@{}c@{}}Insensitive-\\Hyperp.?\end{tabular}}}  & \\
			& & & & & & \\ \midrule
		\multicolumn{7}{l}{\cellcolor[HTML]{EFEFEF}\textbf{(a) VFL-Specific Defenses}} \\
		\midrule
		VFLMonitor~\cite{xuVFLMonitorDefendingOneParty2025} & \xmark & \cmark & \cmark & \cmark & \xmark & \xmark \\
		VFLIP~\cite{choVFLIPBackdoorDefense2024} & \xmark & \cmark & \xmark & \cmark & \cmark & \xmark \\
		GBD~\cite{yuanGeneralTestTimeBackdoor2025} & \xmark & \cmark & \xmark & \cmark & \cmark & \xmark \\
		UBD~\cite{chenUniversalBackdoorDefense2025} & \xmark & \xmark & \cmark & \xmark & \cmark & \xmark \\
		\midrule
		\multicolumn{7}{l}{\cellcolor[HTML]{EFEFEF}\textbf{(b) Generic Defenses}} \\
		\midrule
		Emb. Norm.~\cite{qiuHijackVerticalFederated2024} & \cmark & \cmark & \cmark & \cmark & \cmark & \xmark \\
		Emb. Drop.~\cite{xuVFLMonitorDefendingOneParty2025} & \cmark & \cmark & \cmark & \cmark & \cmark & \xmark \\
		Grad. Noise~\cite{liuTargetedPoisoningAttacks2025} & \cmark & \cmark & \cmark & \cmark & \xmark & \xmark \\
		Grad. Prun.~\cite{chenBackdoorAttackEncryptionProtected2025} & \cmark & \cmark & \cmark & \cmark & \cmark & \xmark \\
		\bottomrule
	\end{tabular}
\end{table}

Backdoor defenses in VFL can be applied at both train time and test time, with three objectives: (i) mitigation, which reduces the attack influence during training; (ii) recovery, which aims to restore the original prediction of attacked samples; and (iii) detection, which identifies attacked samples. We systematize existing works through the lens of the defender's knowledge, defense design, and the overlooked issue of post-attack utility recovery. 

\subsection{Threat Model}

\subsubsection{Defender's Goal}\label{sec:tm-def-goal}
The defender, acting as the active party, aims to prevent backdoor attacks from malicious passive parties and preserve clean model utility simultaneously. This can be achieved in three forms: (i) train-time mitigation; (ii) test-time recovery; and (iii) test-time detection. 

\subsubsection{Defender's Capabilities}\label{sec:tm-def-cap}
Defenders' capabilities are consistent with those of the active party. During training, the defender can inspect, manipulate and store embeddings from all passive parties before feeding them into the top model. The gradients derived from the top model can also be arbitrarily manipulated before sending back to passive parties. During inference, the defender can also manipulate embeddings and the top model.

\subsubsection{Defender's Knowledge}\label{sec:tm-def-knowledge}
VFL-specific defenses require some prior knowledge: (i) clean reference embeddings and (ii) knowledge of the adversarial environment such as the number of attackers and target classes, as summarized in Table~\ref{tab:tax-defense}(a) (Columns 2--3). Obviously, these assumptions are unlikely to hold, and hence we redefine a practical threat model for VFL backdoor defenses as follows:

\begin{figure}[h!]\centering\vspace{-0.5em}
	\begin{tikzpicture}
		\node[draw, fill=gray!10, rectangle, rounded corners, inner sep=4pt] (box) {
			\begin{minipage}{0.97\linewidth}
				\textbf{Practical Threat Model:} The defender possessing the capabilities as an active party (\S\ref{sec:tm-def-cap}) should achieve the defense goal (\S\ref{sec:tm-def-goal}) without clean reference embeddings and the adversarial environment (\S\ref{sec:tm-def-knowledge}).
			\end{minipage}
		};
	\end{tikzpicture}\vspace{-0.5em}
\end{figure}

\noindent Under this threat model, no existing VFL-specific defenses remain practical, as they all explicitly or implicitly rely on clean reference embeddings. 



\subsection{Defense Design}

\subsubsection{Proactive Mitigation}
Our empirical analysis in Section~\ref{sec:experiments-defense} shows the clear advantage of VFL-specific defenses over generic ones, yet only a few methods adopt a proactive approach (Column 4, Table~\ref{tab:tax-defense}). Proactive mitigation before inference deserves more attention, as it reduces the inference burden and better preserves efficiency.

\subsubsection{Multi-target Compatibility}
Similar to attacks, defenses should also be designed to handle multiple target classes. Most current defenses can be adapted (Column 5, Table~\ref{tab:tax-defense}), but UBD~\cite{chenUniversalBackdoorDefense2025} is incompatible, as it assumes a single target class and finetunes the top model to only reduce the attack effect on that class. 

\begin{figure}
	\centering
	\includegraphics[width=\linewidth]{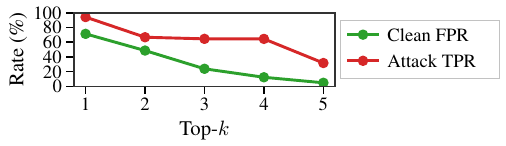}
	\caption{Current defenses are highly sensitive to their hyperparameters. For instance, the top-$k$ class predictions used for detection in VFLMonitor~\cite{xuVFLMonitorDefendingOneParty2025} can drastically affect its effectiveness yet cannot be tuned in practice.}
	\label{fig:vflmonitor_sensitivity_k}
\end{figure}

\subsubsection{Hyperparameter Sensitivity}
A practical defense should be robust to unseen tasks and should not require hyperparameter tuning, so it should be designed to be insensitive to hyperparameter choices. However, some defenses cannot meet this requirement (Column 6, Table~\ref{tab:tax-defense}). For example, VFLMonitor~\cite{xuVFLMonitorDefendingOneParty2025} requires dataset-specific tuning on its $k$ value for top-$k$ predictions of each party. As shown in Figure~\ref{fig:vflmonitor_sensitivity_k}, this value can significantly affect the trade-off between detection performance (Attack TPR) and false alarm rate (Clean FPR) on a VFL dataset. Gradient noise also suffers from the same problem, as different datasets might have varying training gradient magnitudes and an unmatched noise level might be either ineffective or overly harmful to benign learning.


\subsection{Post-attack Utility Recovery}
Current works focus on reducing ASR while overlooking post-attack utility (Column 7, Table~\ref{tab:tax-defense}). In fact, even though some defenses claim to recover clean predictions from trigger-injected ones (e.g., VFLIP~\cite{choVFLIPBackdoorDefense2024}, VFLMonitor~\cite{xuVFLMonitorDefendingOneParty2025}, and UBD~\cite{chenUniversalBackdoorDefense2025}), they all fail miserably to preserve recovery utility. In other words, the attack may be suppressed while predictions remain incorrect. As widely agreed in general backdoor literature (e.g. Robust Accuracy in BackdoorBench~\cite{wu2022backdoorbench}), the defense can be considered failed if the recovery attempt cannot restore the correct prediction.

\begin{table*}
	\setlength\tabcolsep{3.7pt}
	\caption{Current evaluations of VFL backdoor methods are fundamentally flawed. Prior work fails to compare against state-of-the-art baselines, adopts inconsistent and highly sensitive configurations, relies on unrealistic datasets, and provides fragmented analysis with limited metrics. The lack of publicly available implementations further undermines reproducibility, making existing results difficult to trust or build upon.}\label{tab:eval-mistakes}
	\begin{tabular}{@{}l|cc|cccc|c|cccc|c@{}}
		\toprule
		\multicolumn{1}{c|}{\multirow{2}{*}[-0.5ex]{\textbf{Method}}} & \multicolumn{2}{c|}{\textbf{Comparison}} & \multicolumn{4}{c|}{\textbf{VFL Configuration}}                               & \multirow{2}{*}[-0.5ex]{\textbf{\begin{tabular}[c]{@{}c@{}}Realistic\\ Dataset\end{tabular}}} & \multicolumn{4}{c|}{\textbf{Critical Evaluation Aspects}}                              & \multirow{2}{*}[-0.5ex]{\textbf{\begin{tabular}[c]{@{}c@{}}Open\\ Source\end{tabular}}} \\ \cmidrule(lr){2-3} \cmidrule(lr){4-7} \cmidrule(lr){9-12}
		\multicolumn{1}{c|}{}                                 & \textbf{Attack}   & \textbf{Defense}  & \textbf{Top-Depth} & \textbf{Bottom-Arch} & \textbf{Opt} & \textbf{LR} &                                                                                       & \textbf{Efficacy} & \textbf{Robustness} & \textbf{Dependency} & \textbf{Stealthiness} &                                                                                  \\ \midrule
		\multicolumn{13}{l}{\cellcolor[HTML]{EFEFEF}\textbf{(a) Attacks}}\\ \midrule
		BackSplitVFL & \hstar\hstar\hstar & \fstar\hstar\hstar & 3 & CNN & Adam & 1e-3 & \xmark & \fstar\fstar\fstar & \fstar\fstar\hstar & \hstar\hstar\hstar & \fstar\fstar\hstar & \CIRCLE \\
		LMP & \fstar\fstar\hstar & \fstar\fstar\fstar & 1 &     Tiny-ResNet & ? & 1e-3 & \xmark & \fstar\fstar\fstar & \fstar\hstar\hstar & \fstar\fstar\hstar & \hstar\hstar\hstar & \Circle \\
		PMP & \fstar\fstar\hstar & \fstar\fstar\fstar & 1 & Tiny-ResNet & ? & 1e-3 & \xmark & \fstar\fstar\fstar & \fstar\hstar\hstar & \fstar\fstar\hstar & \hstar\hstar\hstar & \Circle \\
		VILLAIN & \fstar\hstar\hstar & \fstar\fstar\hstar & 3 &     VGG16 & SGD & 1e-2 & \xmark & \fstar\fstar\fstar & \fstar\fstar\hstar & \fstar\fstar\fstar & \fstar\hstar\hstar & \LEFTcircle \\
		BadVFL* & \fstar\hstar\hstar & \fstar\hstar\hstar & 1 &     ResNet18 & SGD & 1e-2 & \xmark & \fstar\fstar\fstar & \fstar\hstar\hstar & \fstar\hstar\hstar & \hstar\hstar\hstar & \CIRCLE \\
		LFBA & \hstar\hstar\hstar & \fstar\hstar\hstar & 3 & ResNet18 & Adam & 1e-3 & \xmark & \fstar\fstar\fstar & \fstar\hstar\hstar & \hstar\hstar\hstar & \hstar\hstar\hstar & \CIRCLE \\
		BadVFL & \hstar\hstar\hstar & \fstar\hstar\hstar & 4 &   ResNet18 & ? & ? & \xmark & \fstar\fstar\fstar & \fstar\fstar\hstar & \fstar\hstar\hstar & \fstar\hstar\hstar & \Circle \\
		HijackVFL & \hstar\hstar\hstar & \fstar\hstar\hstar & 3 &   ResNet18 & Adam & 1e-3 & \xmark & \fstar\fstar\fstar & \fstar\hstar\hstar & \fstar\hstar\hstar & \hstar\hstar\hstar & \CIRCLE \\
		BAEVFL & \fstar\fstar\hstar & \fstar\hstar\hstar & 4 & ResNet20 & SGD & 1e-1 & \xmark & \fstar\fstar\fstar & \fstar\fstar\hstar & \hstar\hstar\hstar & \hstar\hstar\hstar & \Circle \\ \midrule
		\multicolumn{13}{l}{\cellcolor[HTML]{EFEFEF}\textbf{(b) Defenses}} \\ \midrule
		\multicolumn{1}{c|}{\multirow{2}{*}[-0.5ex]{\textbf{Method}}} & \multicolumn{2}{c|}{\textbf{Comparison}} & \multicolumn{4}{c|}{\textbf{VFL Configuration}}                               & \multirow{2}{*}[-0.5ex]{\textbf{\begin{tabular}[c]{@{}c@{}}Realistic\\ Dataset\end{tabular}}} & \multicolumn{4}{c|}{\textbf{Critical Evaluation Aspects}}                              & \multirow{2}{*}[-0.5ex]{\textbf{\begin{tabular}[c]{@{}c@{}}Open\\ Source\end{tabular}}} \\ \cmidrule(lr){2-3} \cmidrule(lr){4-7} \cmidrule(lr){9-12}
		\multicolumn{1}{c|}{}                                 & \textbf{Attack}   & \textbf{Defense}  & \textbf{Top-Depth} & \textbf{Bottom-Arch} & \textbf{Opt} & \textbf{LR} &                                                                                       & \textbf{Efficacy} & \textbf{Robustness} & \textbf{Dependency} & \textbf{Recovery} &                                                                                  \\ \midrule
		VFLIP &  \fstar\hstar\hstar & \fstar\fstar\hstar & 3 &     VGG19 & SGD & ? & \xmark & \fstar\fstar\fstar & \fstar\hstar\hstar & \fstar\hstar\hstar & \hstar\hstar\hstar & \LEFTcircle \\
		VFLMonitor & \fstar\hstar\hstar & \fstar\hstar\hstar & 2 & ResNet & Adam & 1e-3 & \xmark & \fstar\fstar\fstar & \fstar\hstar\hstar & \fstar\hstar\hstar & \hstar\hstar\hstar & \Circle \\
		UBD & \fstar\hstar\hstar & \fstar\hstar\hstar & 1 &    ResNet18 & ? & ? & \xmark & \fstar\fstar\fstar & \fstar\fstar\hstar & \fstar\fstar\hstar & \hstar\hstar\hstar & \Circle  \\ 
		GBD & \fstar\fstar\hstar & \fstar\fstar\fstar & 4 & ResNet20 & SGD & 1e-1 & \xmark & \fstar\fstar\fstar & \fstar\hstar\hstar & \fstar\fstar\hstar & \hstar\hstar\hstar & \CIRCLE \\
		\bottomrule
	\end{tabular}\vspace{0.3em}
	{\textbf{[Annotations]} ? = not mentioned in the paper or source code $\vert$ \LEFTcircle = open-sourced but lack key components}
\end{table*}

\section{\scheme{}}\label{sec:benchmark}
After identifying the practicality gap in methodological design, a natural question arises: \emph{how do current attacks and defenses perform in practice?} Unfortunately, the answer remains unclear because the current evaluation practices fail to provide a reliable understanding of backdoor vulnerabilities. A meaningful evaluation should demonstrate improvements over state-of-the-art methods under fair, practical, and comprehensive settings. However, existing studies fall short along all these dimensions, as summarized in Table~\ref{tab:eval-mistakes}.
\begin{itemize}[leftmargin=*, noitemsep, topsep=0pt]
	\item \textbf{State-of-the-art.} Most works compare against only a few baselines (Columns 2--3), partly because reimplementing prior methods requires significant engineering effort, especially when more than half provide no official code (Column 13). As a result, comparisons are incomplete, making it difficult to assess true progress. 
	\item \textbf{Fairness.} Existing evaluations adopt heterogeneous configurations (Columns 4--7), such as different model architectures and hyperparameters. We find that both attacks and defenses are highly sensitive to these choices. Even minor changes, such as model depth or optimizer, can significantly affect performance and favor certain methods. Consequently, reported improvements may reflect configuration bias rather than methodological advances.
	\item \textbf{Practicality.} Existing studies rely on artificial datasets that do not reflect real-world VFL deployments (Column 8). Given the distinct data characteristics of VFL, such simplifications can lead to misleading conclusions. Our analysis shows that methods performing well on synthetic setups can fail under realistic conditions.
	\item \textbf{Comprehensiveness.} Evaluations are fragmented, focusing on a limited set of metrics while overlooking other critical aspects (Columns 9--12). Inconsistent evaluation criteria further hinder meaningful comparison, obscuring the real-world implications of proposed methods.
\end{itemize}

We attribute these shortcomings to the lack of backdoor-focused benchmarks. Existing VFL benchmarks primarily emphasize system performance and provide limited support for evaluating backdoor attacks and defenses (Table~\ref{tab:existing-benchmarks}). Building such a benchmark is challenging: it requires reimplementing diverse methods within a unified framework while ensuring extensibility for future work.
\begin{table}
	\centering
	\setlength\tabcolsep{5.5pt}
	\caption{Existing VFL benchmarks are not designed to evaluate backdoor vulnerabilities. While they support VFL engines and, in some cases, realistic datasets, they provide limited or no support for backdoor vulnerability assessment.}\label{tab:existing-benchmarks}
	\begin{tabular}{@{}lccccc@{}}
		\toprule
		\multirow{2}{*}{\textbf{Benchmark}} & \multirow{2}{*}{\textbf{\begin{tabular}[c]{@{}c@{}}VFL\\ Engine\end{tabular}}} & \multirow{2}{*}{\textbf{\begin{tabular}[c]{@{}c@{}}Realistic\\ Dataset\end{tabular}}} & \multicolumn{3}{c}{\textbf{Backdoor Vulnerability}}      \\ \cmidrule(l){4-6} 
		&                                                                                &                                                                                       & \textbf{Attack} & \textbf{Defense} & \textbf{Analysis} \\ \midrule
		VertiBench~\cite{wuVertiBenchAdvancingFeature2024} &                           \cmark                                                     &    \cmark                                                                                   &    \hstar\hstar\hstar             &   \hstar\hstar\hstar               &  \hstar\hstar\hstar                   \\
		MARS-VFL~\cite{shenMARSVFLUnifiedBenchmark2025} &      \cmark                                                                          &    \cmark                                                                                   &    \fstar\hstar\hstar             &    \fstar\hstar\hstar              &         \fstar\hstar\hstar            \\ 
		VFLAIR~\cite{zouVFLAIRResearchLibrary2023} &          \cmark                                                                      &       \xmark                                                                                &   \fstar\hstar\hstar              &     \fstar\fstar\hstar             &    \fstar\fstar\hstar                 \\ \midrule
		\scheme{} (Ours) &        \cmark                                                                        &     \cmark                                                                                  &    \fstar\fstar\fstar
		& \fstar\fstar\fstar
		&    \fstar\fstar\fstar
		\\ \bottomrule
	\end{tabular}
\end{table}

\begin{figure}
	\centering
	\includegraphics[width=\linewidth]{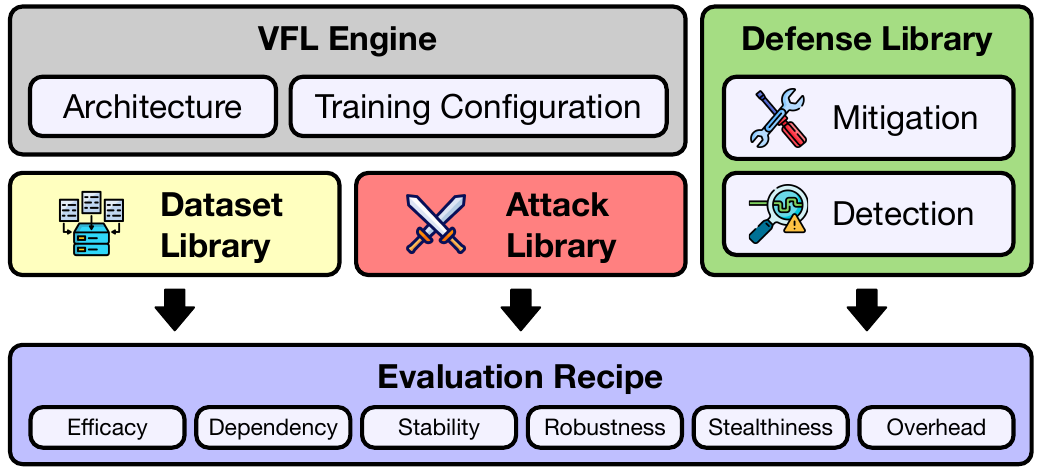}
	\caption{\scheme{} is a modular benchmark for evaluating backdoor vulnerabilities in VFL. It integrates a unified VFL engine, realistic datasets, and standardized implementations of attacks and defenses into a unified evaluation pipeline, enabling fair, practical, and comprehensive assessment.}
	\label{fig:benchmark-flowchart}
\end{figure}

To address these challenges, we introduce \scheme{}, a backdoor-focused benchmark designed to enable fair, practical, and comprehensive evaluation of VFL vulnerabilities, preloaded with state-of-the-art attacks and defenses (Figure~\ref{fig:benchmark-flowchart}). Beyond serving as a toolkit, it provides a standardized evaluation framework for future research. As demonstrated in Section~\ref{sec:experiments-attack} and Section~\ref{sec:experiments-defense}, \scheme{} reveals insights that remain hidden under current evaluation practices. 

\subsection{Ingredients}
\subsubsection{Unified VFL Engine}
We design a unified VFL engine that standardizes model architectures, training configurations, and communication protocols. The engine strictly enforces the information asymmetry inherent in VFL, ensuring that each party accesses only the information consistent with its role. It is extensible to different VFL implementations and supports flexible configurations, enabling controlled variation of architectural and optimization choices.

\subsubsection{Realistic VFL Datasets}
We incorporate realistic VFL datasets (Table~\ref{tab:benchmark-datasets}) that capture practical feature distributions and cross-party characteristics. The benchmark is designed to be easily extensible, allowing new datasets to be integrated with minimal effort. We also include CIFAR-10 as a reference for validating reimplementations, since existing methods all adopt this unrealistic dataset for evaluation, providing a common reference point for comparison.
\begin{table} 
	\centering 
	\setlength\tabcolsep{3.pt}
	\caption{Commonly used datasets such as CIFAR-10 (Row 1) rely on artificial feature splits, whereas the datasets in \scheme{} capture natural cross-party data characteristics.}\label{tab:benchmark-datasets}
	\begin{tabular}{@{}lccccc@{}}
		\toprule
		\textbf{Dataset} & \textbf{Modality} & \textbf{Realistic} & \textbf{\#Parties} & \textbf{\#Classes} & \textbf{\#Samples} \\ \midrule
		CIFAR-10~\cite{krizhevsky2009learning}         & Image             &        \xmark{}           & 2                  & 10                 & 60,000             \\ 
		\midrule
		Satellite~\cite{wuVertiBenchAdvancingFeature2024}        & Image             &        \cmark{}           & 16                 & 4                  & 3,927              \\
		KUHAR~\cite{shenMARSVFLUnifiedBenchmark2025}            & Tabular           &       \cmark{}            & 2                  & 18                 & 20,750             \\
		PTB-XL~\cite{shenMARSVFLUnifiedBenchmark2025}           & Multi-modal       &       \cmark{}            & 3                  & 5                  & 16,244             \\ 
		Vehicle~\cite{wuVertiBenchAdvancingFeature2024}          & Tabular           &       \cmark{}            & 2                  & 3                  & 78,823             \\ 
		NUSWIDE~\cite{wuVertiBenchAdvancingFeature2024}         & Tabular           &    \cmark{}               & 5                  & 2                  & 269,648            \\
		\bottomrule
	\end{tabular}
\end{table}

\subsubsection{State-of-the-art Attacks and Defenses}
We curate a comprehensive collection of state-of-the-art attacks and defenses from major venues. The initial set includes nine attacks and eight defenses, all implemented in PyTorch with over 20,000 lines of code. While some methods provide open-source implementations and can be easily adapted to \scheme{}, others require reimplementation from scratch. For methods whose reproduced performance does not match the originally reported results, we have contacted the authors for clarification and assistance. The framework is designed for extensibility, allowing new methods to be integrated with minimal code changes.

\subsection{Evaluation Recipes}
\scheme{} provides standardized evaluation recipes for attacks and defenses. The recipes define the evaluation aspects and associated metrics that should be reported to enable fair, practical, and reproducible comparison. Detailed formulations are deferred to the appendix.

\subsubsection{Attack Evaluation}
A high attack success rate alone is insufficient to establish a practical attack. In realistic deployments, attacks must also preserve model utility, operate under limited prior knowledge, remain stealthy, generalize across settings, and avoid excessive overhead. We therefore evaluate attacks along the following dimensions.
\begin{itemize}[leftmargin=*, noitemsep, topsep=0pt]
	\item \textbf{Efficacy:} Evaluate attacks using both main task accuracy (MTA) and attack success rate (ASR) across diverse datasets. Since aggregate metrics can mask heterogeneity, conduct per-class analysis to examine variations across target classes and multi-target analysis to assess effectiveness when multiple classes are attacked simultaneously.
	\item \textbf{Dependency:} Quantify how attack effectiveness changes under different levels of required prior knowledge, such as imperfect label knowledge, noisy label inference, or other attack-specific prerequisites, and measure their impact on both ASR and MTA.
	\item \textbf{Stability:} Assess the consistency of attack performance across random seeds and training epochs, rather than reporting only the final checkpoint, using statistics such as the mean and variance of MTA and ASR.
	\item \textbf{Robustness:} Examine attack performance under different VFL environments, including varying numbers of participating parties, and multiple attackers with either shared or conflicting objectives.
	\item \textbf{Stealthiness:} Assess the distinguishability between poisoned and benign embeddings. We introduce stealthiness metrics that capture both abnormal embedding magnitudes and deviations from benign embedding distributions. We consider both an oracle setting based on class-conditional distributions and a label-free setting based on global embedding distributions, and quantify the separability between benign and malicious samples using ROC-AUC.
	\item \textbf{Overhead:} Report the computational, communication, memory, and optimization costs incurred by the attack.
\end{itemize}

\subsubsection{Defense Evaluation}
Reducing attack success alone is insufficient to establish a practical defense. Defenses must preserve normal utility, remain robust across attacks and datasets, and avoid excessive cost. We therefore evaluate defenses along the following dimensions.
\begin{itemize}[leftmargin=*, noitemsep, topsep=0pt]
	\item \textbf{Efficacy:} Evaluate defenses using MTA, defended ASR, and utility recovery (UR) across diverse datasets and attacks. Per-class and multi-target analyses should further assess whether protection remains balanced across target classes and attack objectives.
	\item \textbf{Dependency:} Assess how defense effectiveness depends on required assumptions or prerequisites, such as clean validation data, trusted reference samples, prior attack knowledge, or defense-specific hyperparameter tuning.
	\item \textbf{Stability:} Measure the variability of MTA, defended ASR, and UR across random seeds and repeated runs to ensure that reported protection is reproducible.
	\item \textbf{Robustness:} Evaluate defenses under diverse adversarial environments, including different numbers of parties and scenarios involving multiple attackers with shared or conflicting targets.
	\item \textbf{Overhead:} Report the computation, communication, memory, and inference costs introduced by the defense.
\end{itemize}

Together, these evaluation recipes provide a standardized and practice-oriented framework for assessing VFL backdoor attacks and defenses.

\begin{table*}
	\centering	\setlength\tabcolsep{5.8pt}
	\caption{\scheme{} compares attacks on the same VFL setting across realistic datasets for fairness and generating practical insights. Current attacks can compromise main task accuracy, and their attack success is not as eye-catching as reported.}
	\label{tab:attack_effectiveness}
	\begin{tabular}{clcccccccccc}
		\toprule
		\multirow{2}{*}{\textbf{Metric}} & \multirow{2}{*}{\textbf{Dataset}} & \multirow{2}{*}{\textbf{No Attack}} & \multicolumn{9}{c}{\textbf{Backdoor Attacks}} \\  \cmidrule(l){4-12} 
		 & &  & BackSplitVFL & BAEVFL & BadVFL & BadVFL* & HijackVFL & LFBA & VILLAIN & LMP & PMP \\
		\midrule
		\multirow{6}{*}{\begin{tabular}[c]{@{}c@{}}Main\\Task\\Accuracy\\(MTA)\end{tabular}} & CIFAR-10 & 76.47 & 76.38 & 75.53 & 76.36 & 76.11 & 76.37 & 76.39 & 71.53 & 76.12 & 76.35 \\
		& Satellite & 66.76 & 57.24 & 66.69 & 53.37 & 65.81 & 54.25 & 62.41 & 61.56 & 65.69 & 65.72 \\
		& KUHAR & 62.93 & 60.62 & 38.15 & 62.33 & 38.98 & 62.91 & 41.34 & 12.43 & 60.37 & 40.45 \\
		& PTB-XL & 43.93 & 40.35 & 42.55 & 41.61 & 40.87 & 42.16 & 39.65 & 26.15 & 42.73 & 40.02 \\
		& Vehicle & 83.45 & 83.57 & 84.19 & 83.66 & 83.67 & 83.78 & 83.75 & 84.27 & 83.71 & 83.64 \\
		& NUSWIDE & 78.42 & 78.08 & 78.14 & 78.19 & 78.00 & 78.09 & 78.75 & 78.23 & 78.69 & 78.01 \\
		\midrule
		\multirow{6}{*}{\begin{tabular}[c]{@{}c@{}}Attack\\Success\\Rate\\(ASR)\end{tabular}} & CIFAR-10 & - & 88.04 & 68.11 & 15.49 & 57.19 & 64.12 & 26.33 & 9.99 & 92.04 & 49.59 \\
		& Satellite & - & 100.00 & 27.52 & 7.92 & 26.23 & 38.93 & 21.08 & 15.10 & 69.35 & 29.84 \\
		& KUHAR & - & 72.11 & 5.56 & 5.31 & 6.76 & 7.82 & 7.86 & 0.04 & 36.36 & 42.46 \\
		& PTB-XL & - & 98.84 & 27.85 & 20.38 & 73.60 & 28.75 & 78.37 & 26.34 & 58.30 & 32.89 \\
		& Vehicle & - & 87.61 & 33.33 & 36.64 & 93.27 & 89.56 & 95.06 & 40.87 & 96.19 & 90.60 \\
		& NUSWIDE & - & 82.04 & 69.95 & 49.86 & 99.15 & 96.30 & 98.36 & 55.27 & 99.97 & 94.30 \\
		\bottomrule
	\end{tabular}
\end{table*}
\section{Empirical Analysis: Attacks}\label{sec:experiments-attack}
We empirically analyze existing attacks using \scheme{}. To the best of our knowledge, this is the first study to compare VFL backdoor attacks under a unified benchmark that enforces fairness, practicality, and comprehensive evaluation. Due to the space limit, we highlight key findings here, which reveal that many conclusions drawn from prior studies fail to generalize once these factors are controlled.

\subsection{Efficacy}\label{sec:attack-efficacy}
The conclusions derived from CIFAR-10, the unrealistic dataset commonly used in prior work, differ substantially from those obtained on realistic VFL datasets. As a prerequisite for qualifying as effective attacks, Table~\ref{tab:attack_effectiveness} reports the class-averaged MTA and ASR of all methods across datasets. Although all attacks preserve MTA on CIFAR-10, they can incur substantial utility degradation on realistic datasets, reducing MTA by as much as 13.39\% on Satellite, 50.50\% on KUHAR, and 17.78\% on PTB-XL. The findings are even more concerning for ASR. First, despite prior studies often reporting ASR exceeding 80\% on CIFAR-10, BadVFL, LFBA, and VILLAIN achieve less than 30\% ASR under our evaluation, while most remaining methods fail to exceed 70\%. This discrepancy stems from the fact that \scheme{} enforces configuration fairness: all methods are evaluated under the same VFL environment rather than under method-specific settings. Second, no existing attack demonstrates consistently strong performance across datasets. The gap between the best and worst cases can be substantial, with ASR varying by as much as 92.39\% for BadVFL$^*$. Collectively, these results suggest that current VFL backdoor attacks are considerably less effective than previously implied. Robust attack efficacy across diverse VFL environments remains largely an open problem.

\begin{figure}
	\centering
	\includegraphics[width=\linewidth]{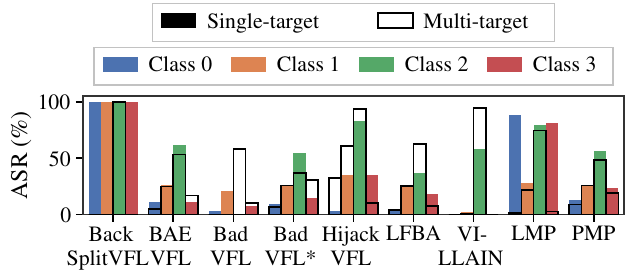}
	\caption{Existing attacks achieve high ASR primarily when targeting the majority class (Class~2), and this bias persists under both single-target and multi-target settings.}
	\label{fig:single_vs_multi_satellite}
\end{figure}

\textbf{Per-Class Analysis.} Existing studies typically report ASR for a selected target class, obscuring variations in attack effectiveness across classes. To expose this phenomenon, we repeat each attack on Satellite using different target classes. Figure~\ref{fig:single_vs_multi_satellite} (color bars) presents the resulting per-class ASR. Almost all attacks achieve their highest ASR when targeting Class~2 (green), which corresponds to the majority class. This observation suggests that attack success is strongly influenced by class distribution and that reported performance can be sensitive to target selection.

\textbf{Multi-Target Analysis.} This tendency becomes even more pronounced in the multi-target setting (hollow bars in Figure~\ref{fig:single_vs_multi_satellite}). Because multi-target learning is inherently more challenging, attacks appear to preferentially optimize shortcuts associated with majority classes when trade-offs arise. Consequently, ASR on minority classes can deteriorate drastically. For example, BackSplitVFL achieves 100\% ASR on the majority class while completely failing on all others. Since multi-target capability is fundamental to the practical backdoor workflow (Section~\ref{sec:attack-design}), improving the balance of effectiveness across targets represents an important direction for future research.

\subsection{Dependency}\label{sec:label-knowledge-dependency}
\begin{figure}
	\centering
	\includegraphics[width=\linewidth]{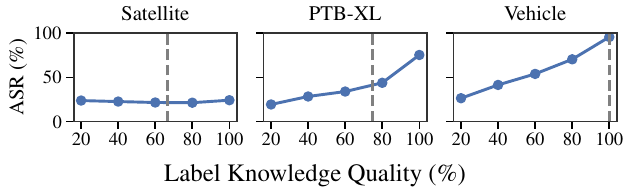}
	\caption{LFBA's attack effectiveness is highly sensitive to label knowledge quality (solid lines), while its label inference component exhibits significant variability across datasets (dashed lines).}
	\label{fig:label_knowledge_dependency_overview}
\end{figure}

Backdoor attacks rely on label knowledge, and the quality of such knowledge can critically influence attack success. Using LFBA, the best-performing attack in our benchmark and one that does not assume perfectly labeled samples as prior knowledge, as a case study, Figure~\ref{fig:label_knowledge_dependency_overview} reports its ASR under different levels of manually controlled label quality (solid lines).
Three observations emerge. First, perfect label knowledge does not necessarily guarantee successful attacks. Even when true labels are provided, LFBA still fails completely on Satellite. Second, on datasets where LFBA is effective (e.g., Vehicle), label quality becomes decisive: reducing label quality by only 20\% can decrease ASR by more than 25\%. Third, the label inference procedure itself exhibits strong dataset dependence. It performs well on some datasets (e.g., Vehicle) but fails on others (e.g., Satellite and PTB-XL), as indicated by the dashed lines. These findings suggest that label knowledge is neither sufficient nor uniformly obtainable in practice. While improving label inference remains important, future work should also investigate why attacks remain ineffective even when accurate label information is available.

\subsection{Stability Under Randomness}\label{sec:attack-reliability}
\begin{figure}
  \centering
    \includegraphics[width=\linewidth]{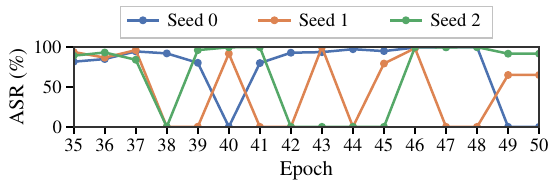}
    \caption{BAEVFL's ASR on Vehicle fluctuates substantially across epochs and is highly sensitive to randomness, highlighting the importance of stability and reproducibility for meaningful attack evaluation.}
    \label{fig:baevfl_vehicle_asr_variance}
\end{figure}

Effective attacks should exhibit consistent performance across repeated executions. However, existing methods can be highly sensitive to randomness. An attack may perform well during one epoch but fail in the next, or succeed under one random seed while collapsing under another. Figure~\ref{fig:baevfl_vehicle_asr_variance} illustrates a representative example using BAEVFL on Vehicle. Its ASR drops from 100\% to 0\% within two consecutive epochs and recovers to 100\% only two epochs later. Changing the random seed introduces further variability. Such instability raises concerns regarding the reliability of current evaluations. An attack may appear successful simply because training happens to terminate at a favorable point along the optimization trajectory. Therefore, future evaluations should systematically assess sensitivity to randomness and demonstrate that attack effectiveness remains stable across both epochs and random seeds.

\subsection{Robustness to Adversarial Environment}\label{sec:attack-generalization}
Backdoor attacks should remain effective regardless of how many peers participate in the VFL system and whether those peers behave benignly or maliciously.

\begin{figure}[!t]
	\centering
	\includegraphics[width=\linewidth]{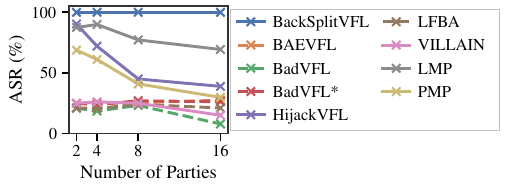}
	\caption{Embedding-space attacks (solid lines) tend to perform better than input-space attacks (dashed lines) on Satellite, no matter the attacker controls more or less features to the top model.}
	\label{fig:satellite_parties}
\end{figure}

The number of peers affects the fraction of top-model inputs controlled by the adversary. Using Satellite as a case study, Figure~\ref{fig:satellite_parties} shows that ASR is strongly influenced by the number of participating parties. Embedding-space attacks (solid lines) consistently outperform input-space attacks (dashed lines), regardless of whether the attacker controls 50\% of the top-model inputs (two parties) or only 6.25\% (sixteen parties). A plausible explanation is that input-space triggers must propagate through the bottom model before influencing the top model, weakening their effect, whereas embedding-space attacks directly manipulate the exchanged representations and bypass this attenuation.

\begin{figure}
	\centering
	\includegraphics[width=\linewidth]{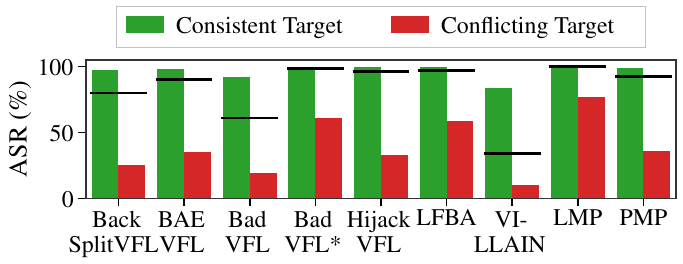}
	\caption{Compared with the single-attacker baseline (solid lines), multiple colluding attackers (green) with consistent targets can reinforce attacks, but independent attackers with conflicting targets (red) can interfere with each other.}
	\label{fig:multi_attacker_nuswide}
\end{figure}

Another overlooked aspect in the adversarial environment concerns the presence of multiple malicious parties. Using NUSWIDE as a case study, Figure~\ref{fig:multi_attacker_nuswide} compares the single-attacker setting (solid lines) with two-attacker settings involving either shared targets (green bars) or conflicting targets (red bars). Across attacks, coordinated attackers targeting the same class substantially improve ASR relative to the single-attacker baseline. In contrast, independent attackers pursuing different targets interfere with one another and significantly degrade attack effectiveness.

These observations suggest that current evaluations underestimate the role of interactions among participants. Understanding how attacks behave under realistic mixtures of benign and adversarial peers remains an important open question.

\subsection{Stealthiness}\label{sec:stealthiness}
\begin{figure}
	\centering
	\includegraphics[width=\linewidth]{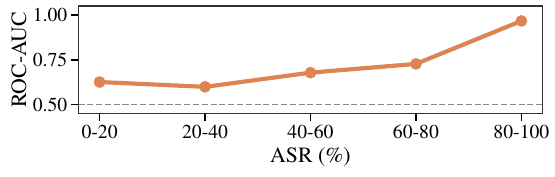}
	\caption{When an attack is effective, the ROC-AUC can be close to 1.0, suggesting a strong separation between clean samples and attack samples. However, when an attack fails, the ROC-AUC can drop to 0.6-0.7.}
	\label{fig:stealthiness_oracle_separability_vs_asr}
\end{figure}

\textbf{Train-Time Stealthiness.} Trigger-injected samples should remain indistinguishable from benign samples throughout training and inference. Although BackSplitVFL and VILLAIN explicitly constrain embeddings to improve stealthiness, the resulting malicious embeddings remain readily distinguishable under the stealthiness score introduced in Section~\ref{sec:benchmark}.
\begin{figure}
	\centering
	\includegraphics[width=\linewidth]{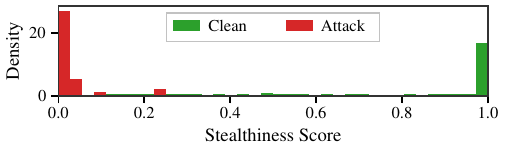}
	\caption{Distribution of stealthiness for clean samples and attack samples. Most clean samples stay in a high level while attacked samples are at a significantly low stealthiness level.}
	\label{fig:stealthiness_distribution}
\end{figure}
Figure~\ref{fig:stealthiness_distribution} illustrates the distribution of stealthiness scores for benign and malicious embeddings generated by BackSplitVFL. The resulting bipolar distribution clearly exposes the presence of poisoning, while the consistently low scores of malicious embeddings can even reveal which samples are poisoned. Consequently, the active party may identify malicious participants during training. This separability is positively correlated with attack effectiveness. Figure~\ref{fig:stealthiness_oracle_separability_vs_asr} reports the ROC-AUC between malicious and benign samples. When ASR is high, the ROC-AUC of stealthiness metrics approaches 1.0, indicating near-perfect detectability. In contrast, when attacks are largely ineffective (ASR $<$20\%), the ROC-AUC drops to approximately 0.6--0.7. Under our simple and intuitive stealthiness metrics, current attacks therefore fail to achieve a favorable balance between efficacy and stealthiness.

\begin{figure}
	\centering
	\includegraphics[width=\linewidth]{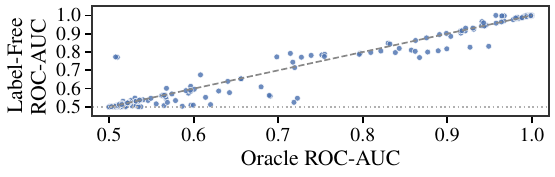}
	\caption{The oracle mode and the label-free mode of our stealthiness metric can both offer strong detectability for effective attacks.}
	\label{fig:stealthiness_label_free_vs_oracle}
\end{figure}

\textbf{Test-Time Stealthiness.} Stealthiness scores can also be leveraged during deployment, where the active party lacks ground-truth labels but seeks to identify suspicious queries. As shown in Figure~\ref{fig:stealthiness_label_free_vs_oracle}, in this label-free setting, we observe strong agreement with the oracle setting that requires labels and is more suitable for training-time detection. In particular, the correlation between label-free and oracle ROC-AUC can reach 0.98. This finding suggests that label-free stealthiness estimation can serve as a practical approximation for both attackers and defenders.

Overall, our results indicate that current VFL backdoor attacks are not stealthy from the defender's perspective. Effective attacks tend to be highly anomalous, whereas stealthier attacks are often ineffective. Whether future methods can achieve a substantially better efficacy-stealthiness trade-off remains an open question. Otherwise, practical VFL backdoor attacks may be considerably easier to detect than previously assumed.

\section{Empirical Analysis: Defenses}\label{sec:experiments-defense}
By systematically comparing VFL backdoor defenses under a unified benchmark, \scheme{} also reveals previously overlooked limitations of existing methods. We summarize the key findings below.
	\begin{table*}
		\centering
		\setlength\tabcolsep{6.5pt}
		\caption{No existing defense can simultaneously preserve benign utility, suppress attacks, and recover correct predictions on attacked samples. ASR, MTA, and UR are averaged over all attacks and target classes.}
		\label{tab:defense_effectiveness}
		\begin{tabular}{clccccccccc}
			\toprule
			\multirow{2}{*}{\textbf{Metric}} & \multirow{2}{*}{\textbf{Dataset}} & \multirow{2}{*}{\textbf{No Defense}} & \multicolumn{4}{c}{\textbf{(a) VFL-specific Backdoor Defenses}}& \multicolumn{4}{c}{\textbf{(b) Generic Backdoor Defenses}} \\  \cmidrule(l){4-7} \cmidrule(l){8-11} 
			& & & VFLIP & VFLMonitor & UBD & GBD & Grad. Prun. & Grad. Noise & Emb. Norm. & Emb. Drop. \\
			\midrule
			\multirow{5}{*}{\begin{tabular}[c]{@{}c@{}}Main\\Task\\Accuracy\\(MTA)\end{tabular}} & Satellite & 63.73 & 18.73 & 56.17 & 66.14 & 60.85 & 60.30 & 40.98 & 54.70 & 68.09 \\
			& KUHAR & 62.81 & 0.31 & 26.17 & 50.58 & 59.98 & 58.87 & 40.94 & 57.53 & 26.96 \\
			& PTB-XL & 41.86 & 5.12 & 25.40 & 39.54 & 40.51 & 35.58 & 14.24 & 42.56 & 46.96 \\
			& Vehicle & 83.70 & 22.44 & 62.52 & 83.13 & 81.10 & 84.69 & 82.71 & 84.66 & 84.57 \\
			& NUSWIDE & 78.08 & 41.94 & 76.74 & 76.86 & 74.02 & 78.71 & 75.93 & 76.73 & 78.65 \\
			\midrule
			\multirow{5}{*}{\begin{tabular}[c]{@{}c@{}}Attack\\Success\\Rate\\(ASR)\end{tabular}} & Satellite & 37.33 & 30.56 & 30.66 & 36.25 & 21.24 & 34.57 & 33.62 & 32.90 & 35.75 \\
			& KUHAR & 20.48 & 6.18 & 7.11 & 21.50 & 15.23 & 20.11 & 15.68 & 13.58 & 17.17 \\
			& PTB-XL & 49.48 & 24.44 & 25.45 & 50.01 & 30.05 & 39.28 & 24.44 & 48.15 & 44.84 \\
			& Vehicle & 73.68 & 33.62 & 29.03 & 72.05 & 37.61 & 67.31 & 57.22 & 73.75 & 75.41 \\
			& NUSWIDE & 82.80 & 50.00 & 60.11 & 72.46 & 71.76 & 79.07 & 54.42 & 54.02 & 78.42 \\
			\midrule
			\multirow{5}{*}{\begin{tabular}[c]{@{}c@{}}Utility\\Recovery\\(UR)\end{tabular}} & Satellite & 48.16 & 17.64 & 49.78 & 41.92 & 43.15 & 47.81 & 32.43 & 44.86 & 49.10 \\
			& KUHAR & 27.20 & 0.32 & 16.18 & 28.45 & 26.01 & 27.32 & 22.16 & 37.15 & 17.92 \\
			& PTB-XL & 27.80 & 6.75 & 18.61 & 23.59 & 26.28 & 27.99 & 13.47 & 29.94 & 34.63 \\
			& Vehicle & 35.66 & 21.30 & 31.79 & 31.56 & 32.06 & 39.68 & 53.69 & 42.50 & 41.99 \\
			& NUSWIDE & 49.36 & 41.94 & 73.24 & 59.03 & 47.85 & 53.22 & 71.56 & 68.31 & 54.84 \\
			\bottomrule
	\end{tabular}
\end{table*}

\subsection{Efficacy}\label{sec:defense-efficacy}
\begin{figure}
	\centering
	\includegraphics[width=\linewidth]{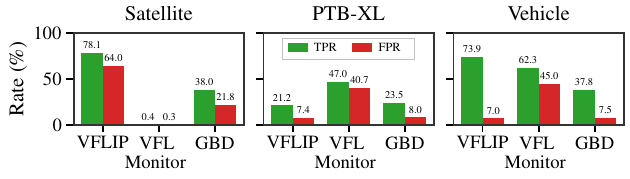}
	\caption{Existing detection-based defenses fail to reliably distinguish attacked samples, preventing them from achieving both high TPR and low FPR across datasets.}
	\label{fig:defense_detection_performance}
\end{figure}

Current defenses fail to simultaneously preserve utility, suppress attacks, and recover correct predictions. Defense efficacy should be evaluated along three dimensions: maintaining main task accuracy (MTA) in the absence of attacks, reducing the attack success rate (ASR) of trigger-injected queries, and achieving utility recovery (UR), i.e., correctly classifying those trigger-injected queries after mitigation. Table~\ref{tab:defense_effectiveness} summarizes the performance of all defenses across datasets, averaged over all attacks and target classes. Three observations emerge. First, existing defenses exhibit a serious utility-security trade-off. Methods that effectively suppress attacks often incur substantial degradation in clean performance, whereas methods that preserve utility typically provide only limited protection. For example, VFLIP achieves the largest ASR reduction on KUHAR, lowering ASR from 20.48\% to 6.18\%, but reduces MTA to only 0.31\%. In contrast, GBD best preserves utility, with MTA decreasing only from 62.81\% to 59.98\%, yet its defended ASR remains at 15.23\%. Similar trade-offs are observed across other defenses and datasets. Second, no existing defense consistently achieves strong performance across datasets. Similar to attacks, defense effectiveness is highly dataset-dependent. A method that performs well in one setting may fail completely in another, suggesting that current defenses do not generalize reliably across realistic VFL environments. Third, existing defenses struggle to recover correct predictions after mitigating attacks. Ideally, UR should approach MTA, indicating that the defense not only suppresses targeted misclassification but also restores the model's intended functionality. However, Table~\ref{tab:defense_effectiveness} shows that no defense achieves this objective, including methods explicitly designed for recovery. For example, although VFLMonitor reduces ASR from 73.68\% to 29.03\% on Vehicle, its UR reaches only 31.79\%, far below its MTA of 62.52\%. Similarly, VFLIP lowers ASR to 33.62\%, yet its UR remains substantially lower than MTA (21.30\% versus 83.70\%). These results indicate that disrupting targeted attacks alone is insufficient, as defended samples often remain misclassified. A plausible explanation is the limited reliability of the anomaly detection mechanisms underlying repair-based defenses. Figure~\ref{fig:defense_detection_performance} examines the detection performance of representative methods. VFLMonitor fails to simultaneously achieve a high true positive rate (TPR) and a low false positive rate (FPR), while VFLIP exhibits strong detection capability on some datasets (e.g., Vehicle) but poor generalization to others (e.g., PTB-XL). Overall, future defenses should place greater emphasis on utility recovery, as preserving prediction correctness is ultimately more important than merely reducing ASR.

\begin{figure}
	\centering
	\includegraphics[width=\linewidth]{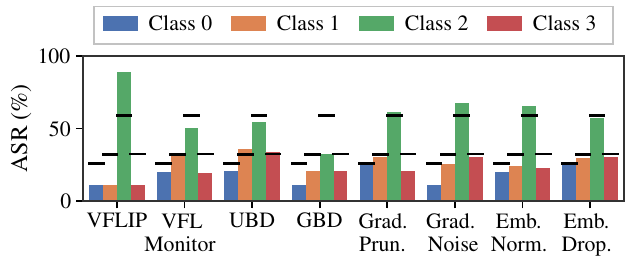}
	\caption{Most defenses fail to provide consistent protection across target classes on Satellite. ASR under defense may even surpass the scenario with no defense (solid lines).}
	\label{fig:defense_seed_variance}
\end{figure}

\textbf{Per-Class Analysis.} Existing studies report class-averaged metrics, which obscure variations in protection across target classes. To expose this phenomenon, Figure~\ref{fig:defense_seed_variance} reports defended ASR for different target classes on the same dataset. Compared with the undefended baseline (solid lines), most defenses fail to consistently suppress attacks across classes. Interestingly, VFLIP reduces the ASR of Classes 0, 1, and 3 to below 10\%, yet strengthens the attack on the majority class (Class~2). Similar trends are observed for gradient noise and embedding normalization. These results suggest that current defenses provide uneven protection and may leave certain classes substantially more vulnerable than others.

\begin{figure}
	\centering
	\includegraphics[width=\linewidth]{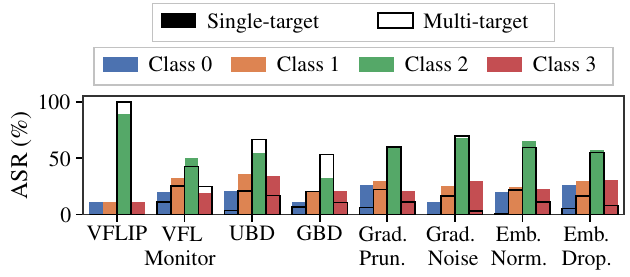}
	\caption{Multi-target attacks further expose the limitations of current defenses: most methods fail to reduce ASR across classes, and the majority-class effect becomes even more pronounced. }
	\label{fig:defense_multi_target_class_asr}
\end{figure}

\textbf{Multi-Target Analysis.} The limitations of current defenses become even more pronounced in the multi-target setting. Figure~\ref{fig:defense_multi_target_class_asr} shows that the defended ASR of most methods remains close to the single-target baseline. Moreover, the influence of class imbalance becomes stronger, with defenses becoming less effective at protecting majority classes from attack. Most defenses exhibit similar robustness under both single-target and multi-target settings despite the latter posing a substantially stronger threat. Since existing methods lack mechanisms for balancing defense effectiveness across classes or attack objectives, improving robustness against multi-target attacks remains an important challenge for future research.

\subsection{Robustness to Adversarial Environment}\label{sec:defense-generalization}
\begin{figure}
  \centering
  \includegraphics[width=\linewidth]{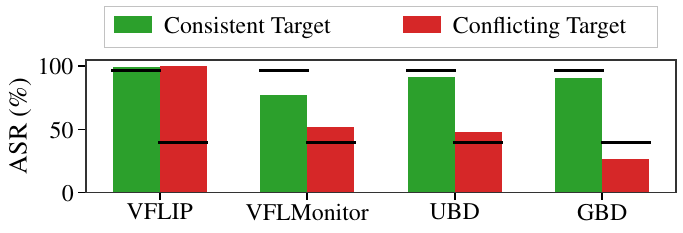}
  \caption{Current defenses are not robust to multi-adversary settings. Compared with the ASR under no defense (solid lines), they provide only marginal protection against colluding attackers and can even inadvertently facilitate attacks when adversaries target different classes.}
  \label{fig:multiattack_overall_joint_defense}
\end{figure}

Current defenses are fragile under realistic multi-adversary environments. Figure~\ref{fig:multiattack_overall_joint_defense} reports defended ASR (bars) and undefended ASR (solid lines) under two scenarios: (i) multiple attackers sharing the same target and (ii) multiple attackers pursuing conflicting targets. We report the ASRs of one of the attackers. Two observations emerge. First, existing defenses provide only limited protection against colluding attackers. Across methods, the defended ASR under shared-target attacks (green bars) remains close to the undefended baseline, indicating that current defenses struggle to mitigate coordinated adversarial behavior. Second, defenses can inadvertently strengthen attacks by redistributing protection unevenly across targets. For example, VFLIP may successfully suppress one adversary's backdoor while simultaneously enabling another adversary to achieve higher attack success. This phenomenon is particularly concerning because it remains hidden when evaluation focuses solely on aggregate metrics. These findings suggest that current defenses are not robust to complex adversarial environments involving multiple malicious participants. Future defense designs should explicitly account for interactions among attackers and ensure balanced protection across target classes and attack objectives.

%

\section{Discussions}\label{sec:discussions}

By systematizing existing work and completing overlooked practicality dimensions through \scheme{}, we show that current understanding of VFL backdoor vulnerability requires substantial revision. Under realistic threat models, standardized evaluation, and comprehensive criteria, most VFL backdoor attacks and defenses are far less practical than previously reported.

On the attack side, no existing attack remains practical, because they all rely on inaccessible task-related knowledge. Even with such knowledge, their effectiveness is still limited by strong dependency, instability, multi-target incompatibility, poor stealthiness, and sensitivity to the VFL environment. Current attacks are hard to justify under realistic assumptions and difficult to sustain in practice.

This does not mean VFL systems are already well protected against vulnerability. Current defenses rely on unreasonable knowledge and fail to consistently achieve all three defense objectives. These limitations worsen in multi-target and multi-adversary settings, where protection can be uneven across classes and fragile to the attack composition.

Our findings carry useful insights for different stakeholders. Future work on attacks and defenses should follow the practical threat model and workflow while meeting practical requirements, such as attack stealthiness and defense utility recovery. Practitioners who want to deploy backdoor-resilient VFL systems should rely on fair and comprehensive evaluation results under \scheme{} rather than heterogeneous reported numbers.

Our study is not exhaustive. We focus on neural-network-based VFL and single-label classification, while real-world VFL may involve other tasks or protocols~\cite{wu2020privacypreservingvfltree,qianTreebasedModelsVertical2025,chen2022verticallyfederatedgraph,wangVFGCNVerticalFederated2025}. Besides, this work aims to recalibrate the understanding of practical VFL backdoor vulnerabilities rather than propose a new practical attack or defense. We hope \scheme{} can serve as a useful tool to steer future VFL backdoor research toward a more practical direction.

\section{Conclusions}\label{sec:conclusions}

Current VFL backdoor research has made significant progress, but its practicality has not been fairly and fully evaluated. We find that the practicality gaps lie not only in methodology designs, but also in evaluation designs. Our findings through systematization and empirical analysis from \scheme{} suggest that current understanding of VFL backdoor vulnerability is incomplete and unreliable, and should be calibrated under the practical threat model and evaluation recipe to guide future research.

\newpage
\bibliographystyle{IEEEtran}
\bibliography{ref}

\appendix

\subsection{Benchmark Setup and Evaluation Details}

\subsubsection{Evaluation Metrics}

We report standardized metrics for both attacks and defenses. Let $\mathcal{Y}=\{1,\ldots,C\}$ denote the label set, let $y_t$ be the target label of a targeted backdoor attack, and let $\mathcal{D}_c$ denote the clean test subset whose ground-truth label is $c$. We write $\mathcal{F}^T(\mathcal{E}_{\boldsymbol{x}})$ for the prediction on a clean sample $\boldsymbol{x}$ and $\mathcal{F}^T(\tilde{\mathcal{E}}_{\boldsymbol{x}})$ for the prediction when the trigger is injected. Considering that realistic VFL datasets are often class-imbalanced, we use macro-averaged metrics by default so that the majority class does not dominate the calculation.

\textbf{Main Task Accuracy (MTA).} We use macro-F1 on the clean test set as the main-task utility metric:
\begin{equation}
\mathrm{MTA} = \frac{1}{C}\sum_{c \in \mathcal{Y}} \mathrm{F1}_c.
\end{equation}
This definition differs from the conventional metric used in the literature. Our goal is to provide a more practical evaluation on model utility across all classes rather than mainly on majority classes.

\textbf{Attack Success Rate (ASR).} We also define ASR as the macro-averaged targeted success rate over all source classes:
\begin{equation}
\mathrm{ASR} = \frac{1}{C}\sum_{c \in \mathcal{Y}}
\frac{1}{|\mathcal{D}_c|}\sum_{\boldsymbol{x} \in \mathcal{D}_c}
\mathbf{1}[\mathcal{F}^T(\tilde{\mathcal{E}}_{\boldsymbol{x}})=t].
\end{equation}
where $t$ is the target label, $\tilde{\mathcal{E}}_{\boldsymbol{x}}$ is the poisoned embedding of sample $\boldsymbol{x}$. We think that this definition prevents biased evaluation on the majority class, as in the conventional definition the attack difficulty on the majority class could significantly skew the overall ASR. Our definition therefore better reflects whether an attack can consistently redirect predictions from \emph{all} classes to the attacker-chosen target.

\textbf{Label Inference Accuracy.} For attacks that require label inference before poisoning, we explicitly denote the inferred label of sample $\boldsymbol{x}$ by $\hat{y}_{\boldsymbol{x}}$. Let $\mathcal{D}^{\mathrm{inf}}$ be the set of samples on which the attacker uses for further poisoning. We evaluate label inference quality using macro-F1, which is aligned with MTA and ASR:
\begin{equation}
\mathrm{LIA} = \frac{1}{C}\sum_{c \in \mathcal{Y}} \mathrm{F1}^{\mathrm{inf}}_c,
\end{equation}
where $\mathrm{F1}^{\mathrm{inf}}_c$ is the F1 score of class $c$ computed from the inferred labels $\{\hat{y}_{\boldsymbol{x}} \mid \boldsymbol{x}\in\mathcal{D}^{\mathrm{inf}}\}$ and the corresponding ground-truth labels $\{y_{\boldsymbol{x}} \mid \boldsymbol{x}\in\mathcal{D}^{\mathrm{inf}}\}$.

\textbf{Stealthiness.} Intuitively, a stealthy attacked sample should still look normal under the clean embedding distribution. We consider a simple metric: the $L_\infty$ norm of the embedding. We measure whether the attack pushes a sample into a region with unusually large embedding norm or distance. To be specific, for a distance $r$, we collect its per-party distribution from training samples, then map each query sample to a scalar value $r(\boldsymbol{x})$ using the CDF of training samples. We think that samples with close-to-one CDF values are more likely to be malicious, while samples whose CDF are small enough are benign. Then the stealthiness score of the distance measure $r$ is then defined as:

\begin{equation}
S(\boldsymbol{x};CDF,r)=\min(1,\, 2(1-CDF(r(\boldsymbol{x})))).
\end{equation}
This score only penalizes samples whose CDF values are more than 0.5. Then we apply this calculation to $L\infty$ norm to form the stealthiness metrics:

\begin{equation}
r_\infty(\boldsymbol{x})=\|\boldsymbol{\mathcal{E}}_{\boldsymbol{x}}\|_\infty
\end{equation}

as the $L_\infty$ norm of the embedding of sample $\boldsymbol{x}$.

We provide two ways to collect the reference distribution and evaluate on test samples: oracle mode and label-free mode. In oracle mode, we assume that labels of test samples are known, and each party builds class-conditional CDFs $CDF_{\infty,c}$ from training samples in class $c$. Therefore, stealthiness scores of test samples can be evaluated from their per-class reference distribution. In label-free mode, we assume that labels of test samples are unknown, and each party builds a single class-agnostic CDF $F_\infty$ from all training samples. In this case, stealthiness scores of test samples are evaluated from the global reference distribution instead.

Overall, oracle mode uses class-conditional reference distributions and test labels, whereas label-free mode uses class-agnostic reference distributions and does not require test labels. For both modes, we report the per-sample stealthiness scores of clean and attack test samples in the form of histograms. We think that the oracle mode provides the most effective evaluation of stealthiness, but in reality test labels are often unavailable, so the label-free mode is more practical.

\textbf{Utility Recovery (UR).} For defenses that aim to recover correct predictions on attacked samples, we report UR as the macro-F1 of the defended model on the attacked test set:
\begin{equation}
	\mathrm{UR} = \frac{1}{C}\sum_{c \in \mathcal{Y}} \mathrm{F1}^{\mathrm{def}}_c,
\end{equation}

where $\mathrm{F1}^{\mathrm{def}}_c$ is the F1 score of class $c$ computed from the defended model's predictions on the attacked test set and the corresponding ground-truth labels. A high UR indicates that the defense can restore the model's intended functionality on attacked samples.

\textbf{Detection Performance. } For defenses that involve test-time detection, we report the true positive rate (TPR) and false positive rate (FPR) of their detection results:
\begin{equation}
\mathrm{TPR} = \frac{\mathrm{TP}}{\mathrm{TP}+\mathrm{FN}},
\qquad
\mathrm{FPR} = \frac{\mathrm{FP}}{\mathrm{FP}+\mathrm{TN}}.
\end{equation}
A high TPR and a low FPR indicate good detection performance, which is an important aspect of defense efficacy.

\subsubsection{Default Configurations}

We summarize the training defaults and the dataset-specific configuration settings used in \scheme{} in Table~\ref{tab:config-shared} and Table~\ref{tab:config-dataset}, respectively. Across all experiments, we keep the same training defaults whenever possible, since in practice, the active party cannot tune the hyperparameters for each dataset. In particular, we use the same training epochs, optimizer with a fixed learning rate and weight decay, and learning-rate schedule across all settings, and reuse the same default VFL architectures for datasets of the same modality whenever possible. Dataset-specific adjustments are restricted to a small set of necessary factors, such as the batch size, loss function, embedding dimension, and the width of the hidden dimensions of the top model, etc.

\begin{table}[t]
    \centering
    \caption{Shared training defaults in \scheme{}.}
    \label{tab:config-shared}
    \begin{tabular}{lc}
        \toprule
        Setting & Value \\
        \midrule
        Epochs & 50 \\
        Optimizer & Adam \\
        Learning rate & 1e-3 \\
        Weight decay & 1e-5 \\
        LR scheduler & Constant \\
        Data augmentation & No \\
        \bottomrule
    \end{tabular}
\end{table}

\begin{table*}[t]
    \centering
    \caption{Dataset-specific scenario configurations in \scheme{}.}
    \label{tab:config-dataset}
    \begin{tabular}{lccccccc}
        \toprule
        Dataset & Modality & \#Parties & Bottom Model & Embedding Dim. & Top Model & BS & Loss \\
        \midrule
        CIFAR-10 & Image & 2 & ResNet18 & 512 & 2-layer MLP (1000) & 128 & CE \\
        Satellite & Image & 16 & ResNet18 & 512 & 2-layer MLP (1000) & 32 & CE \\
        Vehicle & Tabular & 2 & 3-layer FNN (100,100) & 100 & 2-layer MLP (200) & 128 & CE \\
        KUHAR & Tabular & 2 & 3-layer FNN (100,100) & 100 & 2-layer MLP (200) & 128 & CE \\
        NUS-WIDE & Tabular & 5 & 3-layer FNN (100,100) & 100 & 2-layer MLP (200) & 128 & BCE \\
        PTB-XL & Sequential / Tabular & 3 & GRU / 3-layer FNN (100,100)$^\dagger$ & 512 / 100 & 2-layer MLP (1000) & 128 & CE \\
        \bottomrule
    \end{tabular}
    
    \vspace{0.3em}
    {\footnotesize $\dagger$ PTB-XL uses heterogeneous bottom models in the default setting.}
\end{table*}

For attacks and defenses, we mainly follow their default settings in original papers. Some methods have different sets of hyperparameter choices for different datasets and they do not cover the datasets in \scheme{}, so we reuse the hyperparameters for CIFAR10 and apply to all datasets. For hyperparameters not explicitly specified in the paper or source code, we perform a small grid search on CIFAR10 to find the best values, and then fix them across all datasets.

\subsubsection{Extension to Multi-Target Setting}

To extend single-target attacks to the multi-target setting, we preserve each method's original design as much as possible and only replace the single target with all classes in the label space. Target-specific prior knowledge is expanded to all target classes without increasing the total number of known labeled samples, and known labels will be proportionally distributed across all classes. This is to simulate realistic difficulties to obtain those labels. For attacks with class-conditioned triggers, this results in one perturbation or trigger module per class. For attacks with fixed class-agnostic triggers, we keep the trigger pattern unchanged and instead assign non-overlapping trigger positions to different classes whenever possible.

\end{document}